\documentclass[journal]{IEEEtran}
\usepackage{times}
\usepackage{epsfig}
\usepackage{graphicx}
\usepackage{amsmath}
\usepackage{amssymb}

\usepackage{times}
\usepackage{epsfig}
\usepackage{graphicx}
\graphicspath{ {./figs/} }
\usepackage{amsmath}
\usepackage{amssymb}
\usepackage{algorithm}
\usepackage{algorithmic}

\usepackage{array}

\usepackage{float}

\def\ie{i.e.}
\def\eg{e.g.}

\def\0{\textbf{0}}
\def\1{\textbf{1}}

\def\x{\boldsymbol{x}}
\def\y{\boldsymbol{y}}
\def\z{\boldsymbol{z}}

\def\u{\boldsymbol{u}}

\def\II{\mathcal{I}}

\def\RR{\mathbb{R}}

\def\C{\mathcal{C}}

\def\G{\mathcal{G}}

\def\J{\mathcal{J}}

\def\M{\mathcal{M}}

\def\U{\mathcal{U}}
\def\T{\mathcal{T}}

\def\X{\mathcal{X}}
\def\Y{\mathcal{Y}}

\usepackage{multirow}
\usepackage{algorithmic}

\usepackage{graphicx,subfigure}

\newcommand{\myparagraph}[1]{\noindent\textbf{#1.}}

\usepackage[pagebackref=true,breaklinks=true,letterpaper=true,colorlinks,bookmarks=false]{hyperref}

\ifCLASSINFOpdf
\else
\fi
\begin{document}
% \title{{  Cluster-guided Asymmetric Contrastive Learning with Cluster Refinement for Unsupervised Person Re-Identification}}

\title{{Cluster-guided Asymmetric Contrastive Learning for Unsupervised Person Re-Identification}}

%\title{{Cluster-guided Contrastive Learning  with\\ Cluster Refinement for Unsupervised \\Person Re-Identification}}


%\author{\IEEEauthorblockN{Mingkun Li, Chun-Guang Li,~and~Jun Guo}
%\IEEEauthorblockA{School of Artificial Intelligence,\\~Beijing University of Posts and Telecommunications\\
%Email:\{mingkun.li,~lichunguang,~guojun\}@bupt.edu.cn}}

\author{Mingkun~Li,~Chun-Guang~Li,~\IEEEmembership{Senior Member,~IEEE,}  
        ~and~Jun~Guo% <-this % stops a space  
        %~Xianbiao~Qi,~and~Jun~Guo% <-this % stops a space  
        \thanks{M. Li, C.-G. Li and J. Guo are with the School of Artificial Intelligence,  Beijing University of Posts and Telecommunications, Beijing, 100876 P.R. China e-mail: \{mingkun.li, lichunguang, guojun\}@bupt.edu.cn.}
        \thanks{Chun-Guang Li is the corresponding author.}
        % <-this % stops a space
\thanks{Manuscript received xx, 2021; revised xx, xxxx.}
}

\markboth{IEEE Transactions on Image Processing,~Vol.~14, No.~8, April~2022}%
{Li \MakeLowercase{\textit{et al.}}: Cluster-guided Asymmetric Contrastive Learning}

\maketitle

\begin{abstract}
Unsupervised person re-identification (Re-ID) aims to match pedestrian images from different camera views in an unsupervised setting. 
Existing methods for unsupervised person Re-ID are usually built upon the pseudo labels from clustering. 
However, the result of clustering depends heavily on the quality of the learned features, which are overwhelmingly dominated by colors in images. In this paper, we attempt to suppress the negative dominating influence of colors to learn more effective features for unsupervised person Re-ID. 
Specifically, we propose a Cluster-guided Asymmetric Contrastive Learning (CACL) approach for unsupervised person Re-ID, in which clustering result is leveraged to guide the feature learning in a properly designed asymmetric contrastive learning framework. In CACL, both instance-level and cluster-level contrastive learning are employed to help the siamese network learn discriminant features with respect to the clustering result within and between different data augmentation views, respectively.
In addition, we also present a cluster refinement method, and validate that the cluster refinement step helps CACL significantly. 
Extensive experiments conducted on three benchmark datasets demonstrate the superior performance of our proposal. 
\end{abstract}


% \fi
\IEEEpeerreviewmaketitle

\section{Introduction}
\label{sec:intro}

\IEEEPARstart{U}{nsupervised} person Re-identification (Re-ID) aims to match pedestrian images from different camera views in %the 
unsupervised setting without demanding massive labelled data, and has attracted increasing attention in computer vision and pattern recognition community in recent years~\cite{Zheng:arXiv16}. The great challenge we %have to 
face in unsupervised person Re-ID is to tackle %the 
heavy variations from different viewpoints, varying illuminations, changing weather conditions, 
cluttered background and etc., without supervision labels.

\begin{figure}[bth]
\begin{center}
{\includegraphics[clip=true,width=0.8\columnwidth]{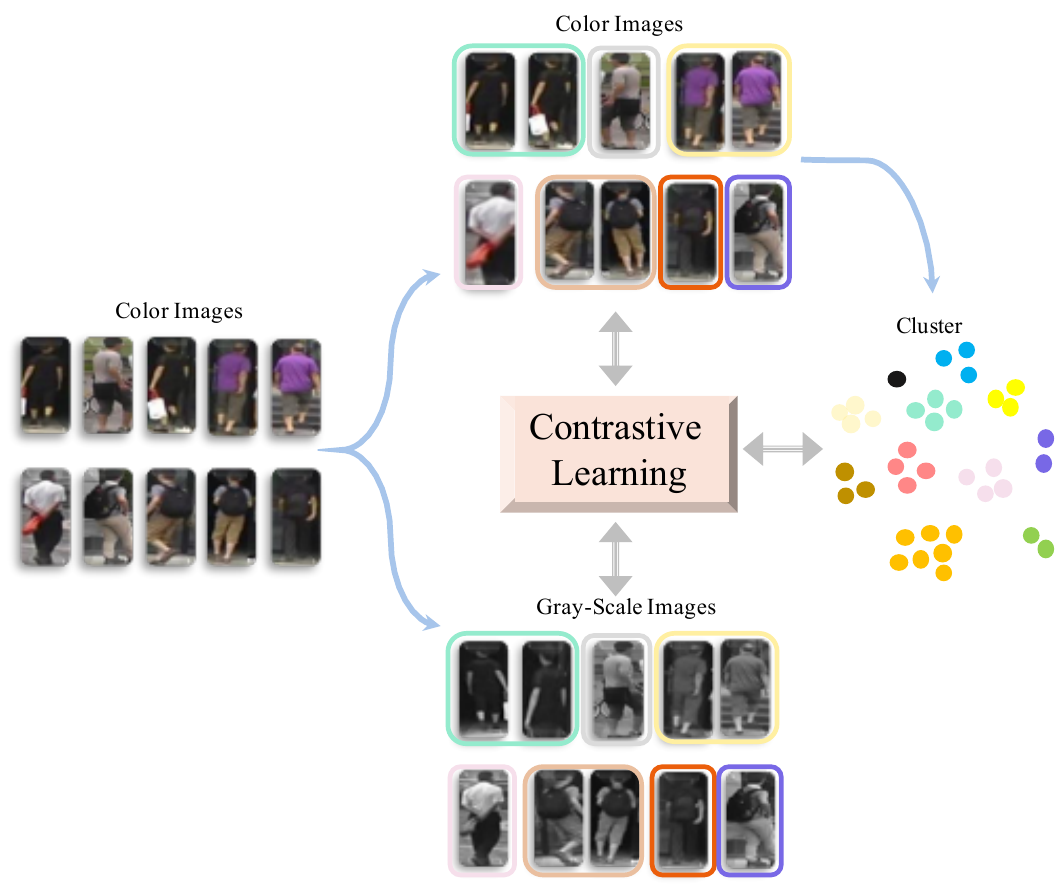}} 
\end{center}
% \caption{The illustration of our method: using a siamese network to find the invariance between the un-color image and the color image, and guiding the training through label sharing between the two type images.}
\caption{Illustration for basic idea of our proposal. We attempt to leverage the clustering information % structure 
into contrastive learning %(via a siamese network) 
to find more effective features %with 
by exploring the invariance between color images and gray-scale images.}

\label{fig:basic-ideas}
\vspace{-8pt}
\end{figure}

Recently, existing methods for unsupervised person Re-ID are usually built on 
exploiting weak supervision information (\eg, pseudo labels) from clustering. 
For example, MMT~\cite{GE:ICLR20} uses DBSCAN~\cite{EsterDBSCAN:AAAI96} algorithm to generate pseudo labels and exploit the pseudo labels to train two networks. HCT~\cite{Zeng:CVPR20} uses a hierarchical clustering algorithm to gradually assign pseudo labels to the training samples during the training stage. SSG~\cite{Fu:ICCV19} uses $k$-means on training samples with multi-views. 
However, the performance of these methods heavily relies on the quality of the pseudo labels, which directly depends on the feature representation of the input images.  

% Here: Mar 20, 2022
More recently, contrastive learning is applied to perform feature learning in unsupervised setting, \eg, \cite{Xie:NIPS20, Caron:arxiv20, Chen:ICLR20, Chen:arxiv20, Larochelle:NIPS2020}. The primary idea in these methods is to learn some invariance in feature representation with self-supervised mechanism based on data augmentation. 
In SimCLR~\cite{Chen:ICLR20}, each sample and its multiple augmentations are treated as positive pairs, and the rest of the samples in the same batch are treated as negative pairs and,  a contrastive loss is used to distinguish the positive and negative samples to prevent the model from falling into a trivial solution. 
We note that SimCLR requires to use a large batch size, \eg, 256 $ \sim $ 4096, to contain enough negative samples for effectively training the networks.  
In BYOL~\cite{Larochelle:NIPS2020} and SimSiam~\cite{Chen:arxiv20}, a predictor layer is used to prevent feature collapse without using negative samples. 
In InterCLR~\cite{Xie:NIPS20} and SwAV~\cite{Caron:arxiv20}, clustering is used to prevent the feature collapse.
In particular, in SwAV~\cite{Caron:arxiv20}, a scalable online clustering loss is proposed to train the siamese network with multi-crop data augmentation; whereas in InterCLR~\cite{Xie:NIPS20}, a MarginNCE loss is proposed to enhance the discriminant power. 
While promising performance has been reported on ImageNet \cite{Krizhevsky:NIPS12}, however, these contrastive learning methods are not suitable for unsupervised person Re-ID due to serious feature collapse. 

In this paper, we attempt to leverage cluster information into contrastive learning to develop an effective framework for unsupervised person Re-ID. We notice that the performance of person Re-ID depends heavily on the effectiveness of the learned features. However, the learned features are overwhelmingly dominated by the colors in pedestrian images (such as the clothing color and background color), especially in the unsupervised setting. For example, the pedestrian images with similar color clothes %similar clothes in similar colors % 
often have smaller distances in feature space, which may result in mistakes in clustering, 
and the mistakes in clustering may further bring wrong guidance to the pseudo labels for training the network.  
Although colors are important feature to match pedestrian images for person Re-ID, it may also become an obstacle to learn more subtle and effective texture features that are important fine-level cues for person Re-ID. Thus it is desirable to learn more robust and discriminating features that can resist dominant colors for person Re-ID task. % In this paper, we attempt to leverage cluster information into contrastive learning to develop an effective framework, as illustrated in Fig.~\ref{fig:basic-ideas}, for unsupervised person Re-ID.

\begin{figure}
\begin{center}
\footnotesize
%\fbox{\rule{0pt}{2in} \rule{.9\linewidth}{0pt}}
\subfigure[Raw ]{\includegraphics[width=0.1250\linewidth]{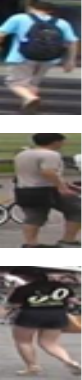}} %
~~~~~~~~
\subfigure[$\T$]{\includegraphics[width=0.125\linewidth]{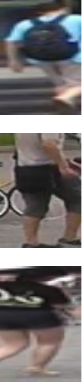}} 
~~~~~~~~
\subfigure[$\T^\prime$]{\includegraphics[width=0.125\linewidth]{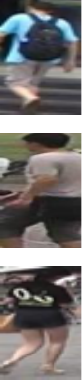}} 
~~~~~~~~
\subfigure[\hspace{-2pt}\footnotesize{$\G\circ\T^\prime$}]{\includegraphics[width=0.125\linewidth]{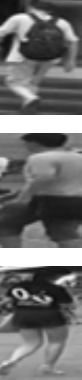}} 

\end{center}
\caption{Illustration for the raw images and the augmented images. The first column shows the raw images. The middle two columns show the images generated with transforms $\T(\cdot)$ and $\T^\prime(\cdot)$. The last column shows the corresponding gray-scale images which  are generated with both transform $\T^\prime(\cdot)$ and color-to-grayscale transform $\G(\cdot)$, \ie, $\G\circ\T^\prime(\cdot)$. }
\label{fig:aug}
% \vspace{-5pt}
\end{figure}

Unfortunately, it is quite challenging to properly suppress the negative impact of colors for learning more effective fine-grain level features without loss of discriminant %color 
information. For example, directly using random color changing (\ie, color-jitter~\cite{chen:cvpr2021}) for data augmentation in contrastive training %will 
may damage the consistency in color distribution, not that helpful to gain generalization ability on unseen samples.  %lose important original color distribution information~\cite{chen:cvpr2021}.}
To this end, in this paper, we propose a novel and effective framework for unsupervised person Re-ID, termed Cluster-guided Asymmetric Contrastive Learning (CACL), in which clustering information is properly incorporated into contrastive learning to learn robust and discriminant features while suppressing dominant colors, as illustrated in Fig.~\ref{fig:basic-ideas}. 
% 
%in which both instance-level contrastive learning and cluster-level contrastive learning are properly conducted to learn robust and discriminant features while suppressing dominant colors. 
To be specific, we explore supervision information from the perspective of suppressing colors in the framework of cluster-guided contrastive learning, in which the samples in asymmetric views of specifically designed data augmentations (\eg, color images vs. gray-scale images) as shown in Fig.~\ref{fig:aug}---are exploited to provide strong supervision to impose invariance in feature learning. 
%that exploits samples in asymmetric views via data augmentation with specifically designed transforms---\eg, color images vs. gray-scale images as shown in Fig.~\ref{fig:aug}---to provide strong supervision which imposes invariance in feature learning. 
By integrating the clustering results into contrastive learning, the proposed framework is able to avoid feature collapse.
By suppressing dominant colors, the proposed framework is able to effectively learn robust and discriminating features other than colors. 
% In addition, in the cluster guided step, we present a cluster refinement method further to improve the quality of the generated raw clusters, thereby helping the proposed contrastive learning to learn more accurately.
In addition, we also present a simple but effective cluster refinement method to improve the clustering result and thus further enhancing the contrastive learning.
We conduct extensive experiments on three benchmark datasets, and experimental results validate the effectiveness of our proposal.
% the proposal's effectiveness. }

\begin{figure*}[!htb]
\begin{center}
\includegraphics[width=0.80\linewidth]{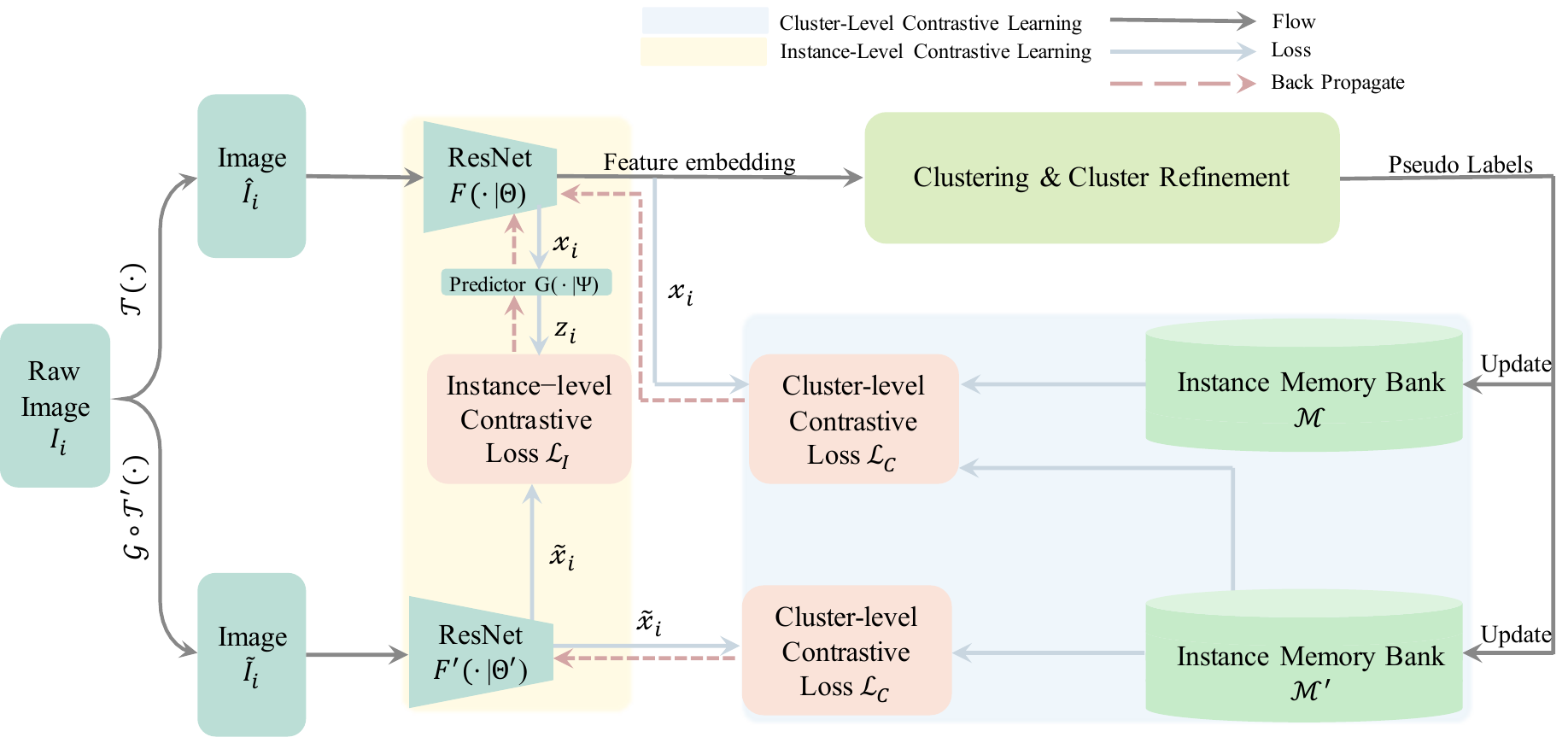}
%\vspace{5pt}
\caption{Illustration for our proposed Cluster-guided Asymmetric Contrastive Learning (CACL) framework. After training, we keep only the ResNet $F(\cdot| \Theta)$ in the first branch for inference and use the feature $\x_i$ for testing.  % {\mcr Propagate --> Propagation}
}
\label{fig:CACL}
 \end{center}
\end{figure*}

% In addition, to further help the model mine cluster-level information more accurately, we propose a method of refining the clustering. Specifically, by filtering out the noise samples contained in the clustering, the clustering is further improved, so that the cluster-cluster-level contrastive learning can more accurately mine out the hidden information in the cluster structure.

\myparagraph{Paper Contributions} 
The contributions of the paper are highlighted as follows. 
\begin{enumerate}

\item %We propose to learn robust and discriminant features other than colors by suppressing colors for unsupervised person Re-ID. To the best of our knowledge, there is no prior work in unsupervised person Re-ID in this line. We propose a Cluster-guided Asymmetric Contrastive Learning (CACL) framework. By suppressing colors, CACL exploits both weak supervision information in clustering { results} and strong supervision information in data augmentation.
% We propose to learn robust and discriminant features other than colors by suppressing colors for unsupervised person Re-ID. In this paper, 
% We propose an effective contrastive learning framework, which is assisted with cluster information, called Cluster-guided Asymmetric Contrastive Learning (CACL). CACL exploits learning fine-grained features by suppressing the negatively dominating influence of the colors in images. To the best of our knowledge, there is no prior work in unsupervised person Re-ID in this line.

We propose an effective unsupervised framework that leverages clustering information into contrastive learning while suppressing the %dominating
 dominant colors in images to learn fine-grained features. 
 
 %To the best of our knowledge, there is no prior work in unsupervised person Re-ID in this line. 

\item  %In CACL, we propose inter-views and intra-view cluster-level contrastive loss functions that can effectively  explore cluster-level hidden information from different data augmentation views.

We propose a novel cluster-level loss function to perform inter-views and intra-view contrastive learning that can effectively exploit the cluster-level hidden information from different data augmentation views.

\item % In the cluster-guided step, we present a cluster refinement method and verify that the refined clusters help contrastive learning framework to improve the performance significantly. 
We also present a cluster refinement method and verify that the refined clustering information helps the contrastive learning framework significantly.

\end{enumerate}

The remainder of this paper is organized as follows. Section \ref{sec:related-work} describes the relevant work. Section \ref{sec:our-proposal} presents our proposal. Section \ref{sec:experiments} shows experiments and Section \ref{sec:conclusion} gives the conclusions.

\section{Related Work}
\label{sec:related-work}
\subsection{Unsupervised Person Re-identification}
Person Re-ID aims to find specific pedestrians from videos or images according to targets. 
For the increasing demand in real life and avoiding the high consumption of labeling datasets, unsupervised person Re-ID has become popular in recent years~\cite{Zheng:arXiv16}.  
The existing unsupervised person Re-ID methods can be divided into two categories: a) unsupervised domain adaptation methods, %which
that need labeled source dataset and unlabeled target dataset~\cite{liu:CVPR19,Bak:ECCV18,Peng:CVPR16,Wang:CVPR18}; and b) pure unsupervised methods, %which 
that need with only unlabeled dataset~\cite{Ge:NIPS20,Lin:AAAI19,Fan:TOMM18,Lin:CVPR20}.

The unsupervised domain adaptation methods train the network with the help of labeled datasets, and transfer the network to unlabeled datasets by reducing 
the gap between two datasets. % \eg, 
For example, \cite{sun:AAAI16} proposed to align the second-order statistics of the distributions in the two domains through linear transformations to reduce the domain shift; \cite{Ge:NIPS20} proposed a combined loss function to co-train with samples from the source and target domains and the merging memory bank; \cite{ganin:arxiv14} proposed to maximize the inter-domain classification loss and minimize the intra-domain classification loss to learn domain robust features. 
However, unsupervised domain adaptation methods are limited by the requirement of the target dataset having close distribution to the source dataset.
% of a small discrepancy between the distributions of the target data and source data.

Most purely unsupervised person Re-ID methods rely on the pseudo labels to train the network. For example, HCT \cite{Zeng:CVPR20} uses hierarchical clustering to generate pseudo labels and train the convolution neural network for feature learning; \cite{Wang:CVPR20} assigns multiple labels to samples and proposes a new loss function for multi-label training. 
Note that the quality of the pseudo labels relies on the feature representation of the input images. However, in the early stage, the feature representation is not good enough to generate high-quality pseudo labels, and thus the low-quality pseudo labels will contaminate the network training. Therefore, it is needed to %we present 
design a cluster refinement method to improve the clustering quality before % generating the pseudo labels.}
feeding the pseudo labels to train the network.

\subsection{Contrastive Learning}

In recent years, with the development and application of the siamese network, contrastive learning began to emerge in the field of unsupervised learning. 
Contrastive learning aims at learning good image representation. It learns invariance in features by 
manipulating a set of positive samples and negative samples with data augmentation.

The existing methods %of 
for contrastive learning can be further categorized to: 
a) instance-level methods~\cite{Chen:ICLR20,Chen:arxiv20,bojanowski:ICML17,dosovitskiy:PAMI15,Larochelle:NIPS2020} and 
b) cluster-level methods~\cite{Caron:arxiv20,Xie:NIPS20,Li:AAAI2021}. 
Instance-level methods regard each image as an individual class and 
consider two augmented views of the same image as positive pairs and treat others in the same batch (or memory bank) as negative pairs. 
For example, SimCLR~\cite{Chen:ICLR20} regards samples in the current batch as the negative samples;  MoCo~\cite{He:CVPR20} uses {a dictionary} 
to implement contrastive learning, which converts one branch of the contrastive learning into a momentum encoder; SimSiam~\cite{Chen:arxiv20} proposed a stop-gradient method that can train the siamese network without negative samples.
Cluster-level methods regard the samples in the same clusters as positive samples %clusters 
and other samples as negative samples. % clusters. ~\cite{Xie:NIPS20}
For example, in \cite{Xie:NIPS20} InfoNCE loss is combined with MarginNCE loss to attract positive samples and repelled negative samples; in~\cite{Caron:arxiv20} multi-crop data augmentation is used to enhance the robustness of the network and a scalable online clustering method is proposed 
to explore the inter-invariance of clusters; in~\cite{Li:AAAI2021} weights-sharing deep neural networks are used to extract features from sample pairs with different data augmentations, and contrastive clustering is performed with respect to both the features in the row and column spaces. 

However, in the unsupervised setting, the instance-level contrastive learning methods simply make each sample independently repel each other, which will undoubtedly ignore the cluster %structure 
information. In contrast, cluster-level contrastive learning can effectively mine cluster %structure 
information, but %cluster-level contrastive learning 
it relies heavily on the clustering result. Unfortunately, %But 
in the early training stage, the features are not good enough to yield good %high quality 
clustering result. % due to the lack of effective guide information to avoid feature collapse. 
Thus, an effective way to train the network by combining %with 
both the two lines of contrastive learning methods is needed.

In this paper, we attempt to bridge the two lines of contrastive learning methods into a unified framework to form effective mutual learning and joint training: a) the instance-level contrastive learning helps training the network to perform 
feature learning---especially in the early training stage; %whereas 
meanwhile b) the cluster-level contrastive learning helps training the network---especially when the quality of the clustering has been improved. In this way, the self-supervision information imposed by data augmentation and the weak supervision information obtained from clustering can be fully exploited without the need to use negative samples pairs.

\section{Our Proposal: Cluster-guided Asymmetric Contrastive Learning (CACL)} 
% with cluster refinement}
\label{sec:our-proposal}

This section  presents our proposal---Cluster-guided Asymmetric Contrastive Learning (CACL) approach for unsupervised person Re-ID. 

%% check to here!!! May 2

For clarity, we show the architecture of our proposed CACL in Fig.~\ref{fig:CACL}. Overall, our CACL is a siamese network, which consists of two branches of backbone networks $F(\cdot|\Theta)$ and $F^\prime(\cdot|\Theta^\prime)$ without sharing parameters, 
where $\Theta$ and $\Theta^\prime$ are the parameters in the two networks, respectively, and a predictor layer $G(\cdot|\Psi)$ is added after the first branch, where $\Psi$ denotes the parameters in the predictor %prediction 
layer. % $G(\cdot)$. 
The backbone networks  $F(\cdot|\Theta)$ and $F^\prime(\cdot|\Theta^\prime)$ are implemented\footnote{It also works if the backbone networks other than ResNet-50 are used.} via ResNet-50~\cite{he:CVPR16resnet} for feature learning.

Given an unlabeled image dataset $\II = {\{I_i}\}_{i=1}^N$ consisting of $N$  samples. 
For an input image $I_i\in \II$, we generate two samples $\hat I_i$ and $\tilde I_i$ via different data augmentation strategies as the inputs of the two branches, respectively, in which $\hat I_i=\T(I_i)$ and $\tilde I_i = \G(\T^\prime(I_i))$,  where $\T(\cdot)$ and $\T^\prime(\cdot)$ denote two different transforms and $\G(\cdot)$ denotes the operation to transform color image into gray-scale image. 
For simplicity, we denote the output features of the first network branch and the second network branch as $\x_i$ and $\tilde \x_i$, and denote the output of the predictor layer in the first branch as $\z_i$, respectively, where $\x_i, \tilde \x_i, \z_i \in \RR^D$.

% We perform clustering with 

The clustering result of the output features $\X:=\{\x_1,\cdots,\x_N\}$ from the first network branch is used to generate the pseudo labels $\Y :=\{\y_1,\cdots,\y_N\}$. We exploit the pseudo labels to leverage the cluster information into the contrastive learning.  % via two set of loss 
% Then we use a refined clustering method to improve the cluster quality and update the pseudo labels. 
%
Specifically, in the training stage, the two network branches $F(\cdot|\Theta)$ and $F^\prime(\cdot|\Theta^\prime)$ are trained with the augmented samples  without sharing parameters, and the pseudo labels  $\Y$ 
are used to guide the training of both network branches. %and also to update the instance memory banks $\M$ and $\tilde \M$, which are initialized with the initial features of the two network branches, respectively.%

% \myparagraph{Instance Memory Banks} 
In CACL, we use instance memory banks $\M = {\{v_i}\}_{i=1}^N$ and $\tilde \M = {\{\tilde v_i}\}_{i=1}^N$ where $v_i, \tilde v_i \in \RR^D$ to store the outputs of two branches, respectively. %repectively,  
Both instance memory banks $\M$ and $\tilde \M$ are initialized with $\X:=\{\x_1,\cdots,\x_N\}$ and $\tilde \X:=\{\tilde \x_1,\cdots,\tilde \x_N\}$, which are the outputs of the network branches $F(\cdot|\Theta)$ and $F^\prime(\cdot|\Theta^\prime)$ pre-trained on ImageNet, respectively.

\subsection{Cluster-guided Contrastive Learning}

 % Density-based 
At beginning, we pre-train the two network branches $F(\cdot|\Theta)$ and $F^\prime(\cdot|\Theta^\prime)$ on ImageNet~\cite{Krizhevsky:NIPS12}, and %employ clustering algorithm  with the features from the first network branch $F(\cdot|\Theta)$ to yield $m$
use the features from the first network branch $F(\cdot|\Theta)$ to yield $m$
clusters, which are denoted as $\C:=\{\C^{(1)}, \C^{(2)}, \cdots, \C^{(m)}\}$.  The clustering result is used to form pseudo labels to train the cluster-guided contrastive learning module.

To exploit the label invariance between the two augmented views and leverage the cluster structure, we employ two types of contrastive losses: a) instance-level contrastive loss, denoted as $\mathcal{L}_{I}$, and b) cluster-level contrastive loss, denoted as $\mathcal{L}_{C}$. 

\myparagraph{Instance-Level Contrastive Loss} 
To match the feature outputs $\z_i$ and $\tilde \x_i$ of the two network branches at instance-level, similar to~\cite{Chen:ICLR20, Larochelle:NIPS2020}, we introduce the negative cosine similarity of the prediction outputs $\z_i$ in the first branch and the feature output of the second branch $\tilde \x_i$ to define an instance-level contrastive loss $\mathcal{L}_{I}$ as follows:
\begin{align}
%\mathcal{L}_I(\z_i,\tilde \x_i) := - \frac{\z^\top_i}{\|\z_i\|_2} \frac{\tilde \x_i}{\| \tilde \x_i \|_2},
\mathcal{L}_I := - \frac{\z^\top_i}{\|\z_i\|_2} \frac{\tilde \x_i}{\| \tilde \x_i \|_2},
\label{eq:Instance-level}
\end{align}
where $\| \cdot \|_2$ is the $\ell_2$-norm.

\myparagraph{Cluster-Level Contrastive Loss} 
To leverage the cluster structure to further explore the hidden information from different views, we propose a cluster-level contrastive loss $\mathcal{L}_{C}$, which is further divided %{ into} 
into inter-views cluster-level contrastive loss and intra-views cluster-level contrastive loss.

\begin{itemize}%这里要不要改为，$\omega(I_i)$
\item Inter-views Cluster-level contrastive loss, denoted as $\mathcal{L}_C^{(inter)}$, which is defined as: 
\begin{align}
\mathcal{L}_C^{(inter)} := -\frac{\z_i^\top}{\|\z_i\|_2} \frac{\tilde \u_{\omega(I_i)}}{\| \tilde \u_{\omega(I_i)} \|_2},
\label{eq:CCL-inter}
\end{align}
where $\omega(I_i)$ is to find the cluster index $\ell$ for $\z_i$, and $\tilde \u_\ell$ is the center vector of the $\ell$-th cluster in which %$\U =\{\u_1,\cdots,\u_{m'}\}$ and 
$\tilde \U :=\{\tilde \u_1,\cdots,\tilde \u_{m'}\}$ and the cluster center $\tilde \u_\ell$ is defined as
\begin{align}
%\omega_{1\ell} = \frac{1}{|\C^{(\ell)}|} \sum_{\x \in {\C^{(\ell)}}} \boldsymbol{m}_{1\x}, 
\tilde \u_\ell = \frac{1}{|\C^{(\ell)}|} \sum_{I_i \in \C^{(\ell)}} \boldsymbol{\tilde v}_i,
\label{eq:cluster_center-tilde}
\end{align}
where $\boldsymbol{\tilde v}_i$ is the instance feature of image $\tilde I_i$ in the instance memory bank $\tilde \M$, {  $\C^{(\ell)}$ is the $\ell$-th cluster.}
The inter-views cluster-level contrastive loss $\mathcal{L}_C^{(inter)}$ %(\x_i) 
defined in Eq.~\eqref{eq:CCL-inter} is used to reduce the discrepancy between the projection output $\z_i$ of the first network branch and the cluster center $\tilde \u_\ell$ of the feature output of the second branch with the gray-scale view.

\item Intra-views Cluster-level contrastive loss, 
denoted as $\mathcal{L}_C^{(intra)}$, which is defined as:
\begin{equation}
\begin{aligned}
%\mathcal{L}_C^{(intra)}(\x_i,\tilde \x_i | \Y ) 
\mathcal{L}_C^{(intra)} 
= & - (1 - q_i)^2 \ln( q_i) \\
  & - (1 - \tilde q_i)^2 \ln(\tilde q_i),
\end{aligned}
\label{eq:CCL-intra}
\end{equation}
where $q_{i}$ and $\tilde q_{i}$ are the softmax of the inner product of the network outputs and the corresponding instance memory bank, which are defined as
\begin{align}
       q_{i}=\frac{\exp(\u_{{\omega(I_i)}}^\top{\x_i}/\tau)}{ \sum_{\ell = 1}^{m'}\exp(\u_{\ell}^\top {\x_i}/\tau)},\\
\tilde q_{i}=\frac{\exp(\tilde \u_{{\omega(I_i)}}^\top{\tilde \x_i}/\tau)}{ \sum_{\ell = 1}^{m'}\exp(\tilde \u_{\ell}^\top {\tilde \x_i}/\tau)},
\label{eq:CCL-q}
\end{align}
%{  
where 
%$\omega(I_i)$ and $\omega(I_i)$ are the pseudo labels of $\x_i$ and $\tilde \x_i$}, and 
$\u_\ell$ and $\tilde \u_\ell$ are the center vectors of the $\ell$-th cluster for the first branch and the second branch, respectively, in which $\tilde \u_\ell$ is defined in Eq.~\eqref{eq:cluster_center-tilde} and $\u_\ell$ is defined as
\begin{align}
\u_\ell = \frac{1}{|\C^{(\ell)}|} \sum_{I_i \in \C^{(\ell)}} \boldsymbol{v}_i, %.
\label{eq:cluster_center}
\end{align}
where %The 
$\boldsymbol{v}_i$ is the instance feature of image $\hat I_i$ in the instance memory bank $\M$. Note that both $\x_i$ and $\tilde \x_i$ share the same pseudo labels $\omega(I_i)$ from clustering.
The intra-views cluster-level contrastive loss $\mathcal{L}_C^{(intra)}$ in Eq.~\eqref{eq:CCL-intra} is used %use 
to encourage the siamese network to learn features with respect to the corresponding cluster center for the two branches, respectively.

\end{itemize}

Putting the loss functions in Eqs.~\eqref{eq:CCL-inter} and \eqref{eq:CCL-intra} together, we have the cluster-level contrastive loss $\mathcal{L}_{C}$ as follows:
\begin{align}
\mathcal{L}_{C} := \mathcal{L}_C^{(inter)} + \mathcal{L}_C^{(intra)}. 
\label{eq:CCL}
\end{align}
%

%Overall, the total loss of our CACL approach can be summarized as % formulate as:
%\begin{align}
%\mathcal{L} = \mathcal{L}_{I} + \mathcal{L}_{C} = \mathcal{L}_{I} + \mathcal{L}_C^{(inter)} + \mathcal{L}_C^{(intra)}. 
%\label{eq:Loss}
%\end{align}

\myparagraph{Remark 1} %Note that 
The cluster-level contrastive loss $\mathcal{L}_{C}$ in Eq. \eqref{eq:CCL} aims to leverage the clustering information % structure 
to minimize the difference between the samples of the same cluster from different augmentation views %(\ie, 
via $\mathcal{L}_C^{(inter)}$, and within the same augmentation view %(\ie, 
via $\mathcal{L}_C^{(intra)}$. This will help the siamese network to  mine the hidden information brought by the basic augmented view in the first branch and the gray-scale augmented view in the second branch to prevent feature collapse to a trivial solution and impose the supervision information to learn features other than colors. 
% { \mc Through  Eq.~\eqref{eq:CCL} the model can learn features of pictures under suppressed colors. }

\subsection{Clustering and Cluster Refinement}
 
%\myparagraph{Cluster Refinement}
% 
Note that the cluster-level contrast loss is greatly affected by the quality of the clustering result. When the clusters are noisy, % contain noises, 
it will cause % have 
negative effects on the training. % loss. 
To improve the quality of the clustering result, %Therefore, 
we propose a cluster %ing 
refinement method which removes a proportion of noisy samples in larger clusters, % clustering result, and 
helping the model to better learn the information at the cluster level.
%
%Basically, we select larger clusters that contain samples %small clusters  with higher intra-similarity and remove lower intra-similarity samples as noisy samples. % from large clusters to refine clusters.

For a cluster, we want to keep the samples with higher similarity and remove the samples with lower similarity.  Given a set of raw clusters, denoted as $\{\C^{(1)}, \C^{(2)}, \cdots, \C^{(m)}\}$, without loss of generality, we pick $\C^{(i)}$ to perform cluster refinement. 
At first, we obtain an over-segmentation of $\C^{(i)}$, \ie, $\C^{(i)}$ is further divided into $\{\C^{(i)}_1,\C^{(i)}_2,\cdots,\C^{(i)}_{n_i}\}$. Then we perform cluster refinement according to the following criterion:
\begin{align}
\begin{aligned}
\text{if}~ D(\C^{(i)}_j|\C^{(i)}) < D(\C^{(i)}) & ,~ \text{then}~ \C^{(i)}_j \text{is kept;}
\end{aligned}
\label{eq:cluster-refinement-criterion}
\end{align}
otherwise $\C^{(i)}_j$ is removed, where $D(\C^{(i)}_j|\C^{(i)})$ is the average inter-distance from all samples in the sub-cluster $\C^{(i)}_j$ to other samples in cluster $\C^{(i)}$, and $D(\C^{(i)})$ is the average intra-distance among samples in cluster $\C^{(i)}$.

After such a post-processing step, the clusters of larger size are improved and at meantime, 
more singletons or tiny clusters are also produced. We denote the refined clusters as $\C^\prime = \{\C^{(1)}, \C^{(2)}, \cdots, \C^{(m')}\}$, where $m' \geq m $. Compared to tiny clusters and singletons, the larger clusters are more informative to provide pseudo supervision information %cluster structure 
to guide the contrastive learning. % important to guide the network  feature learning training. 

\myparagraph{Remark 2}
In implementation, we use DBSCAN algorithm~\cite{EsterDBSCAN:AAAI96} % with Jaccard distance~\cite{zhong:jacarddistance} 
to generate the raw clusters and to generate the over-segmentation of the clusters. 
DBSCAN~\cite{EsterDBSCAN:AAAI96} is a density-based clustering algorithm. It regards a data point as \emph{density reachable} if the data point lies within a small distance threshold $d$ to other %neighboring 
samples, where the parameter $d$ is the distance threshold to find neighboring point. 
Specifically, to generate the raw clusters, we employ DBSCAN with a slightly larger distance threshold parameter $d$ (\eg, $d=0.6$); whereas to generate the over-segmentation, we use a slightly smaller distance threshold parameter $d'$, where $d' := d- \delta$ (\eg, $\delta=0.02$). 
We will show the influence of the parameters $\delta$ and $d$ in %the 
experiments.  
%}

\subsection{Training Procedure for Our CACL Approach}

% \subsection{The Back Propagate in Our Method}

In CACL, the two branches in the siamese network are %is 
implemented with ResNet-50 \cite{he:CVPR16resnet} and they are not %do not 
sharing parameters. We pre-train the two network branches on ImageNet at first and use the learned features to initialize the two memory banks $\M$ and $\tilde \M$, respectively. 

In  training stage, we train both network branches at the same time with the total loss:
\begin{align}
\mathcal{L} := \mathcal{L}_{I} + \mathcal{L}_{C}. 
\label{eq:Loss}
\end{align}

% and because we don't use negative samples in contrastive learning, 
We update the two instance memory banks $\M$ and $\tilde \M$, respectively, as follows:
\begin{align}
\boldsymbol{v}_i^{(t)} \leftarrow \alpha \boldsymbol{v}_i^{(t-1)} + (1-\alpha) \x_i,
\label{eq:update-cluster_1}\\
\boldsymbol{\tilde v}_i^{(t)} \leftarrow \alpha \boldsymbol{\tilde v}_i^{(t-1)} + (1-\alpha)\tilde \x_i,
\label{eq:update-cluster_2}
\end{align}
where  $\alpha$ is set as 0.2 by default (and we will discuss the influence of  $\alpha$ in experiments).

%To prevent collapse
In order to save the computation cost\footnote{Note that it is not necessary to use the stop-gradient operation in our CACL because the clustering result provides enough guide information under the asymmetric structure to prevent collapse. Although this is similar to the method in SimSiam~\cite{Chen:arxiv20}, %mentioned, 
the purpose is different and it is not necessary to use in our proposal.}, %not the same, and it does not have to be used in our model.}, 
we also use a stop-gradient operation as mentioned in SimSiam~\cite{Chen:arxiv20}.
Note that we adopt the stop-gradient operation~\cite{Chen:arxiv20} to the second network branch $F^\prime(\cdot|\Theta^\prime)$ when using the instance level loss $\mathcal{L}_I$ in Eq.~\eqref{eq:Instance-level} to perform back propagation. 
Thus, the parameters $\Theta^\prime$ in the second network branch are updated only with the intra-views cluster-level contrastive loss $\mathcal{L}_C^{(intra)}$ in Eq.~\eqref{eq:CCL-intra}.

% For clarity, we summarize the details of the training procedure  in Algorithm \ref{alg:Framwork}.

\myparagraph{Remark 3} For clarity, we summarize the details of the training procedure in Algorithm \ref{alg:Framwork}. 
We note %that the proposed framework for cluster-guided contrastive learning is asymmetry in following three aspects: 
that the ``\emph{asymmetry}'' in the proposed framework for cluster-guided contrastive learning lies %is 
in following three aspects:
a) asymmetry in network structure, \ie, a predictor layer is only added after the first branch\footnote{ It is also feasible to add another predictor layer after the second branch to have a symmetric network structure. Nevertheless, our experimental results show that merely marginal performance improvement can be yielded after adding an extra predictor layer. Thus, we prefer to use the asymmetric network architecture for the contrastive learning framework.}; and b) asymmetry in data augmentation, \ie, the augmented samples provided to the second branch are further transformed into gray-scale; c) asymmetry in pseudo labels generation, \ie, the output features of the first branch are used to generate pseudo labels which are shared with the second branch.  %lables and the cluster result is shared by tow branches.  
Because of the asymmetry in the three aspects mentioned above, we term the proposed framework as Cluster-guided \emph{Asymmetric} Contrastive Learning (CACL).  
% 

%\myparagraph{Comparison to Prior Work} 
\myparagraph{Remark 4} There have been many unsupervised Re-ID methods~\cite{Ge:NIPS20,liu:CVPR19,zhai:CVPR20,chen:cvpr2021} used the contrastive learning to learn %discriminate 
discriminant features. Most of them~\cite{liu:CVPR19,zhai:CVPR20,chen:cvpr2021} are Generative Adversarial Networks (GANs)-based methods and need additional supervised information to assist the training. 
For example, ATNet~\cite{liu:CVPR19} trains multiple %GANS 
GANs through utilizing illumination and camera information, GCL~\cite{chen:cvpr2021} introduces the pose information in training, and AD-cluster~\cite{zhai:CVPR20} uses generating cross-camera samples to assist the training. % 
Unlike these methods, our proposed CACL uses an asymmetric Siamese network to effectively learn fine-grained features by suppressing color with simple data augmentation operations during the training, rather than using an expensive sample generation via GANs.  
Compared to GANs based methods, our CACL is simple, efficient and effective.

\subsection{Inference Procedure for CACL}
After training, we keep only the ResNet $F(\cdot|\Theta)$ in the first branch for inference in testing. 

To be specific, in the inference procedure, we use the output features $\X$ of the first branch $F(\cdot|\Theta)$ to calculate the similarity between images. Given the query image dataset $\II^g = {\{I^g_i}\}_{i=1}^{N^g}$ and the query image dataset $\II^q = {\{I^q_i}\}_{i=1}^{N^q}$, where $N^g$ and $N^q$ are the sizes of the two datasets, respectively. 
For each image $I^g_i$ in the query, we compute the distances between the query image and the images in the gallery $\II^q$ via the feature obtained from the output of the first branch. And then, we sort the distance in ascending order %and 
to find the matched images.

\begin{algorithm}[tb] % [htbp]
	
		\caption{Training Procedure for CACL}
			\label{alg:Framwork}
		\begin{algorithmic}[1]
		\REQUIRE
		 Given a dataset $\II = \{ I_i \}_{i=1}^N$. % 
		
		\ENSURE
		\STATE Pre-train the two network branches on ImageNet.
		\STATE Initialize the two instance memory banks $\M$ and $\tilde M$ and set  $P = P_{best}=0$.
		
		\WHILE{epoch $\leq$ total epoch}
		\STATE Generate $\hat I_i$ and $\tilde I_i$ via data augmentation $\T(\cdot)$ and $\G(\T^\prime(\cdot))$;
		\STATE Perform feature extraction to get $\x_i$ and $\tilde \x_i$;
		\STATE Perform clustering and clustering refinement via Eq.~\eqref{eq:cluster-refinement-criterion} to yield pseudo label $\Y = \{ \y_1, \cdots, \y_N\}$; 
		\STATE Update the two cluster centers $\U$ and $\tilde \U$ via Eq.~\eqref{eq:cluster_center};
		\STATE Train siamese network, \ie, updatomg $\Theta$, $\Psi$ and $\Theta^\prime$ via the total loss in Eq.~\eqref{eq:Loss};
		\STATE Update instance memory bank $\M$ and $\tilde\M$ via Eq.~\eqref{eq:update-cluster_1} and  Eq.~\eqref{eq:update-cluster_2};
		\STATE Evaluate the model performance $P$ with $F(\cdot|\Theta)$;
		\IF{ $P > P_{best}$}	
			 \STATE Output the best model $F(\cdot|\Theta)$ and set $P_{best} \leftarrow P$;
		\ENDIF		
		\ENDWHILE
		\end{algorithmic}

	\end{algorithm}

\section{Experiments}
\label{sec:experiments}

% To validate the effectiveness of our proposal, we conduct extensive experiments on three benchmark datasets. 
In this section, we describe %introduce 
the used benchmark datasets %we use and introduce 
and the detailed parameter settings in experiments at first, and then %of our method. We show 
provide extensive experiments %the performance of our method 
on these datasets, including % and conduct 
a set of detailed ablation study %on each component of our method to show 
and a set of evaluation experiments to show the effect of each component. Finally, we give a set of data visualization experiments. \footnote{The code can be downloaded from \url{https://github.com/MingkunLishigure/CACL}.}
%\footnote{The code can be downloaded from \url{https://github.com/MingkunLishigure/CACL}.}
%have done more analysis on our method and compared it with similar methods.
%
%

\begin{table*}
% \small
\caption{Comparison to other state-of-the-art methods. 'UDA' is to refer the unsupervised domain adaptation methods and 'US' is to refer the purely unsupervised learning methods. { '*'  means % the backbone is  ??? per-trained on ImageNet.}}
that the used backbone is pre-trained on ImageNet.
}}
\begin{center}
\begin{tabular}{|c|c|c|c|c|c|c|c|c|c|c|c|}
\hline
	\multirow{2}{*}{Method} & 
	\multirow{2}{*}{Type}&
					\multirow{2}{*}{Reference} & \multirow{2}{*}{Bakcbone} &
					\multicolumn{4}{c|}{Market-1501} &
					\multicolumn{4}{c|}{DukeMTMC-ReID} \\

	\cline{5-12}
					& & & & mAP & Rank-1 &Rank-5&Rank-10 & mAP & Rank-1 &Rank-5&Rank-10 \\
\hline\hline
%\multirow{14}{*}{UDA}
					PTGAN~\cite{Wei:CVPR18}&UDA&CVPR'18&GoogleNet~\cite{googlenet} &15.7 &38.6&57.3&- &13.5 &27.4&43.6&-\\
					SPGAN~\cite{Deng:CVPR18}&UDA& CVPR'18  &{ResNet50*~\cite{he:CVPR16resnet}}&26.7&58.1&76.0&82.7&26.4&46.9&62.6&68.5\\
TJ-AIDL~\cite{Wang:CVPR18}&UDA&CVPR'18&{MobileNet*~\cite{howard:2017Mobilenets}} &26.5&58.2&74.8&-&23.0&44.3&59.6&-\\
 PGPPM~\cite{Yang:arXiv18}&UDA &CVPR'18&{ResNet50*~\cite{he:CVPR16resnet}}&33.9&63.9&81.1 &86.4 &17.9&36.3&54.0 &61.6\\
HHL~\cite{Zhong:ECCV18}&UDA& ECCV'18& {ResNet50*~\cite{he:CVPR16resnet}}& 31.4&62.2 & 78.0 &84.0& 27.2&46.9 & 61.0 &66.7\\

					%MAR & CVPR19  & {\bf48.0} & {\bf67.1} &{\bf79.8} & - \\

					SSG~\cite{Fu:ICCV19} &UDA& ECCV'19&{ResNet50*~\cite{he:CVPR16resnet}}& 58.3&80.0 & 90.0 &92.4& 53.4&73.0& 80.6 &83.2\\
					AD-cluster~\cite{zhai:CVPR20} &UDA& CVPR'20&{ResNet50*~\cite{he:CVPR16resnet}}& 68.3&86.7 & 94.4 &96.5& 54.1&72.6& 82.5 &85.5\\
					ADTC~\cite{ji:ECCV20}&UDA & ECCV'20&{ResNet50*~\cite{he:CVPR16resnet}}& 59.7&79.3 & 90.8 &94.1& 52.5&71.9& 84.1 &87.5\\
					MMCL~\cite{Wang:CVPR20}&UDA & CVPR'20&{ResNet50*~\cite{he:CVPR16resnet}}& 60.4&84.4 & 92.8 &95.0& 51.4&72.4& 82.9 &85.0\\
					MMT~\cite{GE:ICLR20} &UDA&ICLR'20&{ResNet50*~\cite{he:CVPR16resnet}}& 73.8& 89.5 & 96.0 &97.6& 62.3&76.3& 87.7 &91.2\\
					JVTC~\cite{Li:ECCV20} &UDA& ECCV'20&{ResNet50*~\cite{he:CVPR16resnet}}& 67.2&86.8 & 95.2 &97.1& 66.5&80.4& 89.9 &93.7\\
					MEB~\cite{zhai:ECCV20}&UDA & ECCV'20&{ResNet50*~\cite{he:CVPR16resnet}}& 76.0&89.9 & 95.2 &96.9& 65.3 &81.2& {  90.9} &92.2\\
					NRMT~\cite{Zhao:ECCV20} &UDA& ECCV'20&{ResNet50*~\cite{he:CVPR16resnet}}& 71.7&87.8 & 94.6 &96.5& 62.2&77.8& 86.9 &89.5\\
					SpCL~\cite{Ge:NIPS20}&UDA&NIPS'20&{ResNet50*~\cite{he:CVPR16resnet}}&{76.7}& {90.3} & {96.2} &{97.7}&{68.8}&{\bf82.9}& 90.1 &{92.5}\\

	\hline\hline
					%\multirow{9}{*}{USL}
					CAMEL~\cite{Yu:ICCV17}&US&ICCV'17 &{ResNet50*~\cite{he:CVPR16resnet}}&26.3 &54.4&73.1&79.6 &19.8 &40.2&57.5&64.9\\
					Bow~\cite{Zheng:ICCV15}&US& ICCV'15&-& 14.8& 35.8 & 52.4& 60.3& 8.5& 17.1 & 28.8& 34.9\\
					PUL~\cite{Fan:TOMM18}&US&TOMM'18&{ResNet50*~\cite{he:CVPR16resnet}}&22.8&51.5&70.1&76.8 &22.3&41.1&46.6&63.0\\
					LOMO~\cite{Liao:CVPR15}&US &CVPR'15&-& 8.0 &27.2 & 41.6 &49.1 & 4.8 &12.3 & 21.3 &26.6\\

					BUC~\cite{Lin:AAAI19}&US& AAAI'19&{ResNet50*~\cite{he:CVPR16resnet}} &30.6 &61.0&71.6&76.4& 21.9&40.2 & 52.7& 57.4\\
				
					HCT~\cite{Zeng:CVPR20}&US& CVPR'20&{ResNet50*~\cite{he:CVPR16resnet}} &56.4 &80.0 &91.6 &95.2 &50.1 &69.6 &83.4 &87.4\\
					SSL~\cite{Lin:CVPR20}&US& CVPR'20&{ResNet50*~\cite{he:CVPR16resnet}} &37.8 &71.7 &83.8 &87.4 &28.6 &52.5 &63.5 &68.9\\
					SpCL~\cite{Ge:NIPS20}&US&NIPS'20&{ResNet50*~\cite{he:CVPR16resnet}}& 73.1& 88.1 & 96.3 &97.7& 65.3&81.2& 90.3 &92.2\\
					CAP~\cite{Wang:AAAI20}&US&AAAI'20&{ResNet50*~\cite{he:CVPR16resnet}}& 79.2& 91.4 & 96.3 &97.7& 67.3&81.1& 89.3 &91.8\\
					\hline\hline
					{\bf CACL }& US & {\bf This paper}&{\bf ResNet50*~\cite{he:CVPR16resnet}}& {\bf80.9}&{\bf 92.7}& {\bf97.4 }&
					{\bf98.5}&{\bf69.6}&{\underline {82.6}}& {\bf91.2}& {\bf93.8} \\
					{\bf CACL }& US & {\bf This paper}&{\bf IBN-ResNet*~\cite{Pan:IBNResNet}}& {\bf83.6}&{\bf93.3}& {\bf97.7 }&
					{\bf98.3}&{\bf72.5}&{\bf{85.5}}& {\bf92.9}& {\bf94.9} \\
% Theirs & Frumpy \\
% Yours & Frobbly \\
% Ours & Makes one's heart Frob\\
\hline
\end{tabular}
\end{center}
\label{tab:SOTA}
% \vspace{-10pt}
\end{table*}

\subsection{Dataset Description}
To evaluate the effectiveness of our proposal, we use the following three benchmark datasets: 
Market-1501~\cite{Zheng:ICCV15}, DukeMTMC-ReID~\cite{Ristani:Duke} and MSMT17~\cite{MSMT:18}. 

{\bf Market-1501} has 32,668 photos of 1501 people from six different camera views. The training set contains %{ contains} 12936 
12,936 of  751 identities. The testing set contains 19,732 images of 750 identities. 

{\bf DukeMTMC-ReID} consists of images sampling from DukeMTMC-ReID video dataset, 120 frames per video, with a total of 36,411 images of people of 1404  identities. %{ The} 
The training set contains 16,522 images of 702 identities and the testing set contains 2228 query images of 702 identities and 17,661 gallery images. These images are taken from eight cameras.

{\bf MSMT17} has a total of 126,441 images under 15 camera views. The training set contains 32,621 images of 1041 identities. The testing set contains 93,820 images of 3060 identities are used for testing. %The 
MSMT17 is larger than Market-1501 and DukeMTMC-ReID.

\subsection{Implementation Details}

\myparagraph{Settings for Training} In our CACL approach, we use ResNet-50~\cite{he:CVPR16resnet} pre-trained on ImageNet~\cite{Krizhevsky:NIPS12} for both  network branches.\footnote{In Section \ref{subsec:comp-to-sota}, we also provide the performance evaluation with other backbone networks for the two branches.}
The feature outputs $\x_i \in \RR^D$ and $\tilde \x_i \in \RR^D$ of the two networks $F(\cdot|\Theta)$ and $F(\cdot|\Theta^\prime)$ are $D$-dimensional vectors where $D=2048$. We use the features output $\x_i$ of the first branch $F(\cdot|\Theta)$ to perform clustering, where $\x_i =F(\hat I_i|\Theta) \in  \RR^D$. 

The prediction layer $G(\cdot)$ is a $D \times D$ full connection layer. 
We initialize the two memory banks with the outputs of the feature from the corresponding network branches $F(\cdot|\Theta)$ and $F^\prime(\cdot|\Theta^\prime)$, respectively. 
We optimize the network through Adam optimizer \cite{Kingma:arXiv2014} with a weight decay of 0.0005 and train the network with 80 epochs in total. The learning rate is initially set as 0.00035 and decreased to one-tenth per 20 epochs. 
The batch size is set to 64. 
The temperature coefficient $\tau$ in Eq.~\eqref{eq:CCL-q} is set to $0.05$ and the update %upgrade 
factor $\alpha$ in Eqs.~\eqref{eq:update-cluster_1} and \eqref{eq:update-cluster_2} is set to $0.2$.
%\footnote{The code will be released when the manuscript is getting accepted.} 
%\footnote{The code can be downloaded from \url{https://github.com/MingkunLishigure/CACL}.}
%In the training on Market-1501 and DukeMTMC-ReID, the Step 6 in Algorithm~1 costs about 50$\sim$60 seconds where the task of calculating pairwise distance takes about 30 seconds and the task of generating pseudo labels with cluster refinement takes 10$\sim$20 seconds.

\myparagraph{Settings for Data Augmentation} 
In our experiments, we use the same data augmentation operations as other methods~\cite{Ge:NIPS20,GE:ICLR20}, including random horizontal flip, random erasing and random crop, to define data augmentation $\T(\cdot)$ and $\T^\prime(\cdot)$. Besides, we add a gray-scale transform to the input of the second branch.

\myparagraph{Metrics for Performance Evaluation} In evaluation, we use the mean average precision (mAP) and cumulative matching characteristic (CMC) at Rank-1, 5, 10 to evaluate the performance. 
% \footnote{{\mcr The code can be downloaded from \url{https://github.com/mingkunli/CACL}}} 

\subsection{Comparison to the State-of-the-art Methods} %With SOTA}
\label{subsec:comp-to-sota}

We compare our proposed CACL to the state-of-the-art unsupervised domain adaptation methods and purely unsupervised methods for person Re-ID. The purely unsupervised methods for person Re-ID include: CAMEL~\cite{Yu:ICCV17}, PUL~\cite{Fan:TOMM18}, SSL~\cite{Lin:CVPR20}, LOMO~\cite{Liao:CVPR15}, BOW~\cite{Zheng:ICCV15}, BUC~\cite{Lin:AAAI19},  HCT~\cite{Zeng:CVPR20}, SpCL~\cite{Ge:NIPS20}, and CAP~\cite{Wang:AAAI20}. The unsupervised domain adaptation methods for person Re-ID include: PTGAN~\cite{Wei:CVPR18}, ADTC~\cite{ji:ECCV20}, HHL~\cite{Zhong:ECCV18},  SSG~\cite{Fu:ICCV19}, MMCL~\cite{Wang:CVPR20}, AD-Cluster~\cite{zhai:CVPR20}, MEB~\cite{zhai:ECCV20}, NRMT~\cite{Zhao:ECCV20}, SPGAN~\cite{Deng:CVPR18}, TJ-AIDL~\cite{Wang:CVPR18}, JVTC~\cite{Li:ECCV20},  PGPPM~\cite{Yang:arXiv18}, and MMT~\cite{GE:ICLR20}.

The comparison results of the state-of-the-art unsupervised domain adaptation methods and purely unsupervised methods are shown in Table~\ref{tab:SOTA}. We can find 
that our proposed CACL 
achieves 80.9/92.7\% at mAP/Rank-1 on Market-1501 and 69.6/82.6\% at mAP/Rank-1 on DukeMTMC-ReID, respectively. It can be found that CACL not only performs better than all pure unsupervised methods but also achieves the best performance than unsupervised domain adaptation methods.

Moreover, we also conduct experiments on a much larger dataset MSMT17 and report the experimental results in Table \ref{tab:17}. Again, we can observe that our proposed CACL  achieves a leading performance, \ie, 23.0/48.4\% at mAP/Rank-1. It is worth to note that our CACL yields superior performance than some UDA methods on this challenging dataset. These results confirm the effectiveness of our proposal.

Note that Instance-Batch Normalization (IBN)~\cite{Pan:IBNResNet} has been used in object recognition %methods 
and has been proved very effective. Here, we evaluate our CACL, in which the backbone is implemented with %on the 
Instance-Batch Normalization ResNet (IBN-ResNet). Similar to %Compare to %with the 
CACL with ResNet~\cite{he:CVPR16resnet}, we introduce an Instance-Batch Normalization (IBN) layer %is introduced 
to replace the BN layer and call it an IBN-ResNet. 
As shown in 
Table~\ref{tab:SOTA}, the performance of our CACL 
%with 
can be further improved when combining with IBN-ResNet.

\begin{table}
\begin{center}
\caption{Experimental Results on MSMT17.}
\setlength{\tabcolsep}{0.6mm}{
\begin{tabular}{|c|c|c|c|c|c|c|}
					\hline		
					\multirow{2}{*}{Method} &
					\multirow{2}{*}{Type} &
					\multirow{2}{*}{Reference} &
					\multicolumn{4}{c|}{MSMT17}\\
					\cline{4-7}  
					& & &mAP & Rank-1 &Rank-5&Rank-10\\
					
\hline\hline
PTGAN~\cite{Wei:CVPR18}&UDA&CVPR'18&3.3&11.8&-&27.4\\
ECN~\cite{Zhong:CVPR19}&UDA&CVPR'19&10.2&30.2&41.5&46.8\\
SSG~\cite{Fu:ICCV19} & UDA& ICCV'19& 13.3&32.2&-&51.2\\
MMCL~\cite{Wang:CVPR20} & UDA& CVPR'20& 16.2&43.6&54.3&58.9\\
JVTC+~\cite{Li:ECCV20}&US &ECCV'20& 17.3&43.1&53.8&59.4\\
SpCL~\cite{Ge:NIPS20} &  US& NIPS'20&19.1&42.3&55.6&61.2\\
MMT~\cite{GE:ICLR20} &UDA& ICLR'20&24.0&50.1&63.5&69.3\\
SpCL~\cite{Ge:NIPS20}&UDA&NIPS'20&26.8&53.7&79.3&83.1\\
\hline\hline
{\bf CACL}   &US & \bf This paper & \bf 23.0& \bf 48.9& \bf 61.2& \bf 66.4 \\
{\bf CACL w/ IBN-ResNet}   &US & \bf This paper & \bf 29.9& \bf 57.1& \bf 68.4& \bf 73.1 \\

\hline

\end{tabular}
}
\label{tab:17}
\end{center}

% \vspace{-12pt}
\end{table}

\begin{table*}
%\small
\begin{center}
\caption{ Ablation Study on Market-1501 and DukeMTMC-ReID.}
\setlength{\tabcolsep}{0.9mm}
{
\begin{tabular}{|c|p{1.7cm}<{\centering} | p{1.0cm}<{\centering} | p{1.0cm}<{\centering} | p{1.0cm}<{\centering}  |c|c|c|c|c|c|c|c|}
					\hline		
					 \multirow{2}{*}{Components}&\multirow{2}{*}{Cluster Refine} & \multirow{2}{*}{$\mathcal{L}_{I}$} &\multirow{2}{*}{$\mathcal{L}_C^{intra}$} & \multirow{2}{*}{$\mathcal{L}_C^{inter}$} &\multicolumn{4}{c|}{Market-1501}&\multicolumn{4}{c|}{DukeMTMC-ReID}\\
					\cline{6-13}  
					&    &   &  &  & mAP & Rank-1 &Rank-5&Rank-10& mAP & Rank-1 &Rank-5&Rank-10\\
					
\hline\hline
Baseline&    &   &  &  &68.1 & 85.2&94.0 &96.0 & 62.5&78.5 & 88.5 & 90.3\\ 

+ $\mathcal{L}_{C}$   &    &   & \checkmark & \checkmark  &70.8	& 87.5& 	94.4& 	96  & 62.5	&79.5&	88.4&	90.8\\

+ $\mathcal{L}_{I}$    &    & \checkmark  &  &  &74.7 &	88.7 &	95	 &96.6 &	64.2 &	80.7 &	89	 &91.6 \\

+ $\mathcal{L}_{I}$ + $\mathcal{L}_{C}$   &    & \checkmark  & \checkmark & \checkmark &74.4 &89.3 &95.9 &96.7& 63.8&79.2&89.2&91.7 \\
+ Cluster Refine&  \checkmark &   &  &  &73	&87.8&	95.7&	97.2&	65.7&	81.1&	90.6&93.2 \\ 
 
+ Cluster Refine + $\mathcal{L}_{I}$&  \checkmark  & \checkmark  &  &  & 78.2&	91.2&	97&	98.1& 67.6	&81.8	&90.2&	93 \\+ Cluster Refine + $\mathcal{L}_{I}$ +$\mathcal{L}_C^{inter}$ &  \checkmark  & \checkmark  &  & \checkmark & 78.7&	91.2&	97&	97.9&68.5&	81.9&	91.2&	93.8 \\   
+ Cluster Refine + $\mathcal{L}_{I}$ +$\mathcal{L}_C^{intra}$ & \checkmark   &  \checkmark & \checkmark &  &79.2&	91.9&	96.7&	98&	68.3&	82.1&	90.3&	93.2 \\

+ Cluster Refine + $\mathcal{L}_{C}$&   \checkmark &   &  \checkmark& \checkmark &80.4 & 92.2  &97.1  &98.2& 68.8&	82.2&	91.3&	93.8\\

Our CACL    & \checkmark   & \checkmark  &  \checkmark&  \checkmark& {\bf80.9}&{\bf92.7}& {\bf97.4 }&{\bf98.5} &{\bf69.6}&{\bf {82.6}}& {\bf91.2}& {\bf93.8}\\

\hline

\end{tabular}}
\label{TAb:ab}
\end{center}
\end{table*}

\subsection{Ablation Study}

To evaluate the effectiveness of each component: $\mathcal{L}_{I}$, $\mathcal{L}_{C}^{(inter)}$, $\mathcal{L}_{C}^{(intra)}$ and clustering with refinement in our CACL approach, we conduct a set of ablation experiments on Market-1501 and DukeMTMC-ReID.

In the baseline method, we train both branches with data augmentation $\T^\prime(\cdot)$ and  $\T^\prime(\cdot)$ by using the Non-Parametric Softmax loss~\cite{Wu1:CVPR18}, which is defined as 
\begin{align}
\mathcal{L}(\x_i) = -\ln(\frac{\exp(\u_{{\omega(I_i)}}^\top{\x_i}/\tau)}{ \sum_{\ell = 1}^{m'}\exp(\u_{\ell}^\top {\x_i}/\tau)}), 
%loss(x) = -(1-(1 - I(\C_{w(x)}))p_x)^{2 + \lambda I(\C_{w(x)})} \ln(p_x) , 
\label{eq:=cross-entropy-loss-function}
\end{align}
and both the training process and the memory updating strategy in the baseline method are kept the same as our CACL method.  %as same as  CACL method.}

To comprehensively evaluate the contribution of each component, we conduct a set of ablation experiments by testing each component in our CACL framework individually, \ie, cluster refinement,  instance-level contrastive loss $\mathcal{L}_{I}$ and cluster-level contrastive loss $\mathcal{L}_{C}$. To further evaluate the sub-part of the cluster-level contrastive loss, we also conduct experiments to evaluate the influence of using $\mathcal{L}_C^{(inter)}$ or $\mathcal{L}_C^{(intra)}$, separately.

In the ablation experiments, to test the model with contrastive loss $\mathcal{L}_{C}$ or $\mathcal{L}_{I}$, we train both branches with data augmentation $\T^\prime(\cdot)$ and $\G(\T^\prime(\cdot))$, respectively. To test the model performance with the cluster-level contrastive loss $\mathcal{L}_C$ and the sub-part of $\mathcal{L}_C^{(intra)}$, compared to the baseline method,  we need to replace the Non-Parametric Softmax loss in Eq.~\eqref{eq:=cross-entropy-loss-function} by the   loss in Eq.~\eqref{eq:CCL-intra} for both branches. The results of the ablation study are reported in Table~\ref{TAb:ab}.  %%%%%%%%%%%%%%%%%%%%%%%%%%%%%%%%%%%%%%%%%%%%%%%%%%%%%%%%%%%%%%%%%%%%%%%%%%

As can be read in Table~\ref{TAb:ab}, the performance improves when each component is used individually. This validates that each component contributes to the performance improvements. For the experiments of using both  $\mathcal{L}_{C}$ and $\mathcal{L}_{I}$, it does not significantly better than just using $\mathcal{L}_{I}$, and in the experiments of using $\mathcal{L}_{C}$ %use 
we observe a slight improvement than the baseline. This is because the clustering result is not high quality and using $\mathcal{L}_{C}$ will make the training %model 
pay more attention to the noisy cluster information. Therefore, it might bring misleading information to the network training. % model. And 
In the experiments of using both $\mathcal{L}_{C}$ and cluster refinement, we observe significant performance improvement than using the cluster refinement alone. This also validates that the cluster refinement improves the clustering result and the refined clustering information can further enhance the effectiveness of using $\mathcal{L}_{C}$ to train the network. % { without much additional computing sources. \footnote{In the training on datasets Market-1501 and DukeMTMC-ReID, in total Step 6 in Algorithm 1 costs about 1 minute: the task of calculating global distance takes about 30 seconds, and the task of generating pseudo labels with cluster refinement takes 10$\sim$20 seconds.}}
%thus helps $\mathcal{L}_{C}$  effectively exploit the hidden information in samples.

\subsection{More Evaluation and Analysis}

%% check at Here May 2: 01:51

\myparagraph{Evaluation on Importance of Cluster-Guided}   We use an instance-level contrastive loss in our method to mine the invariance between different augment views based on  SimSiam~\cite{Chen:arxiv20}. %However, the model can not be trained effectively by just using contrast learning without the clustering method. To prove this, we train our method without cluster guidance and just train with the instance-level contrastive loss in Eq.~\eqref{eq:Instance-level}. The results are shown in Table~\ref{Tab:Ablation Study_inSupportingMaterial}.
To verify whether the clustering guidance is vital in the contrast learning framework, we train our CACL framework but just using the instance-level contrastive loss in Eq.~\eqref{eq:Instance-level} without the clustering guidance. 
The experimental results are shown in Table~\ref{Tab:Ablation Study_inSupportingMaterial}.
As can be read from Table~\ref{Tab:Ablation Study_inSupportingMaterial}, surprisingly, the contrastive learning framework without clustering guidance did not work at all.

\begin{table}[!ht]
\small
\caption{Ablation Study on Market-1501.}
\begin{center}
{
\begin{tabular}{|c|c|c|c|c|}
					\hline		
					\multirow{2}{*}{Components} & \multicolumn{4}{c|}{Market-1501}\\
					\cline{2-5}  
					& mAP & Rank-1 &Rank-5&Rank-10\\
					
\hline\hline
{ CACL w/o clustering}  & 0.3 &0.5 &1.2 &2.3 \\
\hline
CACL w/o  $stopGrad$ & 80.2 &92.0 &97.0 &97.6\\
CACL  & {80.9}&{92.7}& {97.4 }&{98.5}\\

\hline
\end{tabular}
}
\end{center}

\label{Tab:Ablation Study_inSupportingMaterial}
\end{table}

\begin{figure}
\begin{center}
%\fbox{\rule{0pt}{2in} \rule{.9\linewidth}{0pt}}
\includegraphics[width=0.85\linewidth]{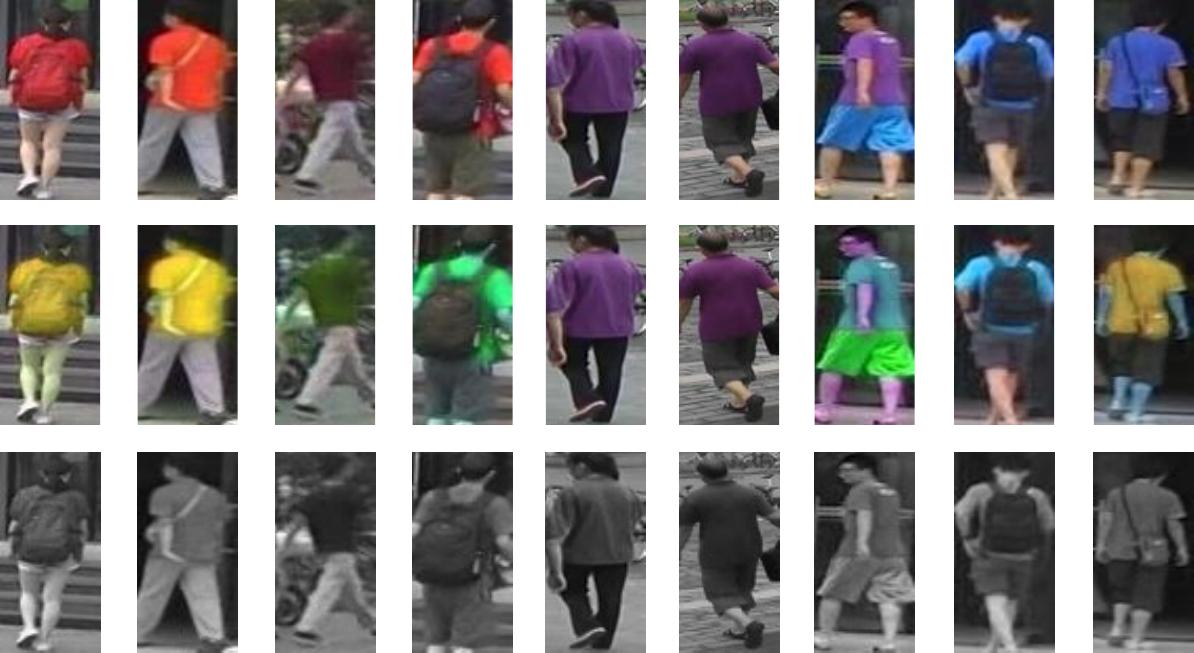} % flowchart.pdf
\end{center}
  \caption{ Illustration for the raw images and the augmented images.  The 1st row: ``raw images''. The 2nd row: ``color-jitter''. Bottom row: ``gray-scale''. }
  
\label{fig:color}
\end{figure}

\myparagraph{Improvements Brought by Suppressing Colors} 
To suppress colors influence,  CACL uses a gray-scale process $\G(\cdot)$ over the data augmentation $\T^\prime(\cdot)$ for the second network branch.  To validate the effectiveness of suppressing colors,  we conduct a set of experiments under different settings: a) simply using data augmentation $\T^\prime(\cdot)$ with raw color; b) using another data augmentation approach, named ``color-jitter'', which denoted as $\J(\cdot)$ to replace $\G(\cdot)$, which output is still a color image; 
c) with gray-scale transform $\G(\cdot)$ after $\T^\prime(\cdot)$.  It should be emphasized that in the implementation, the ``color-jitter'' operation will give random amplitude values to the image changing.
We display the image samples processed with different data augmentation methods in Fig.~\ref{fig:color}. As can be observed, ``color-jitter'' did change the image, but the color information still dominates.

\begin{figure*}[!ht]
\begin{center}
\footnotesize
%\fbox{\rule{0pt}{2in} \rule{.9\linewidth}{0pt}}
\subfigure[Raw Images ]{\includegraphics[trim={0cm 0cm 0cm 0cm},clip,width=0.67\columnwidth]{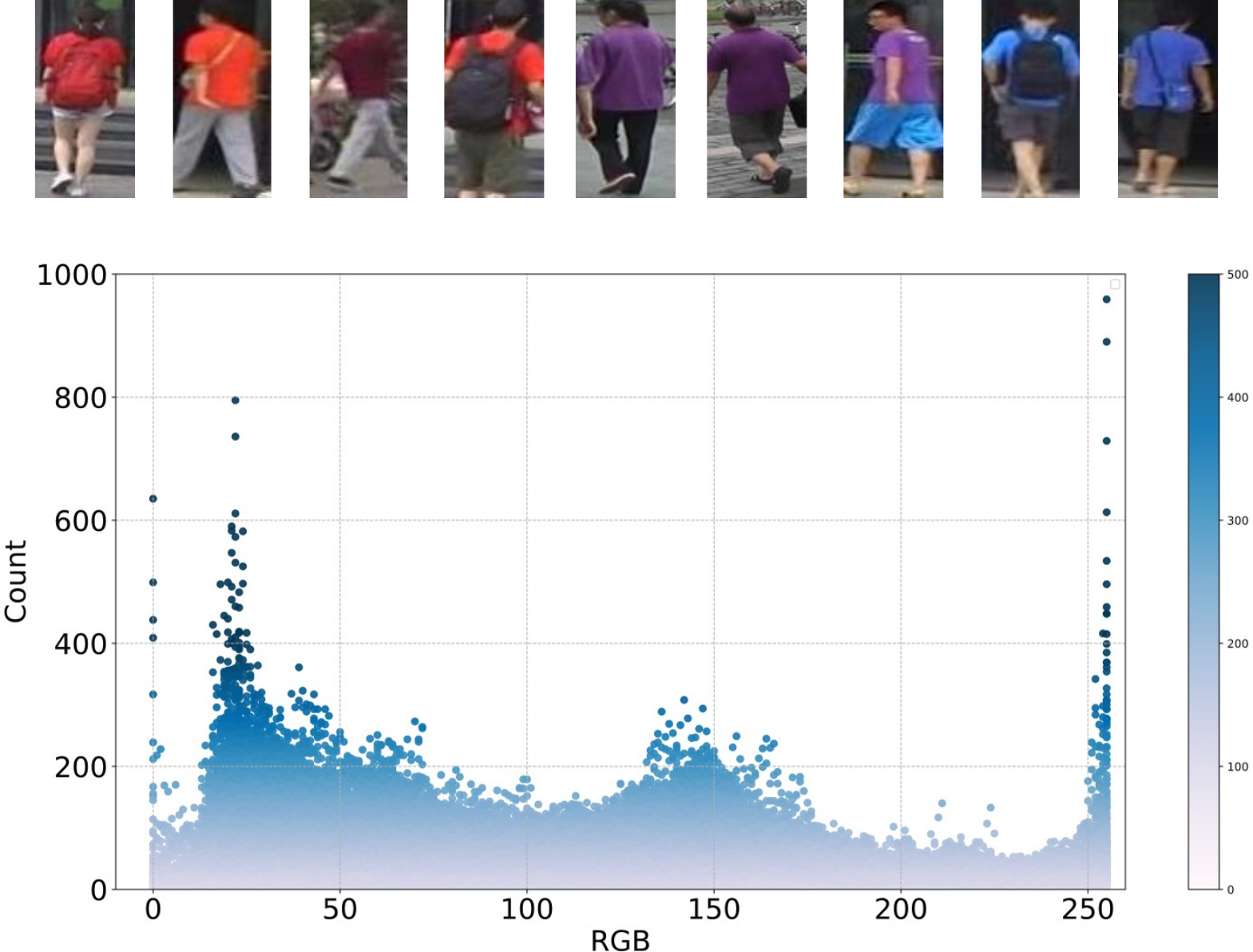}} %
\subfigure[Color-Jitter]{\includegraphics[trim={0cm 0cm 0cm 0cm},clip,width=0.67\columnwidth]{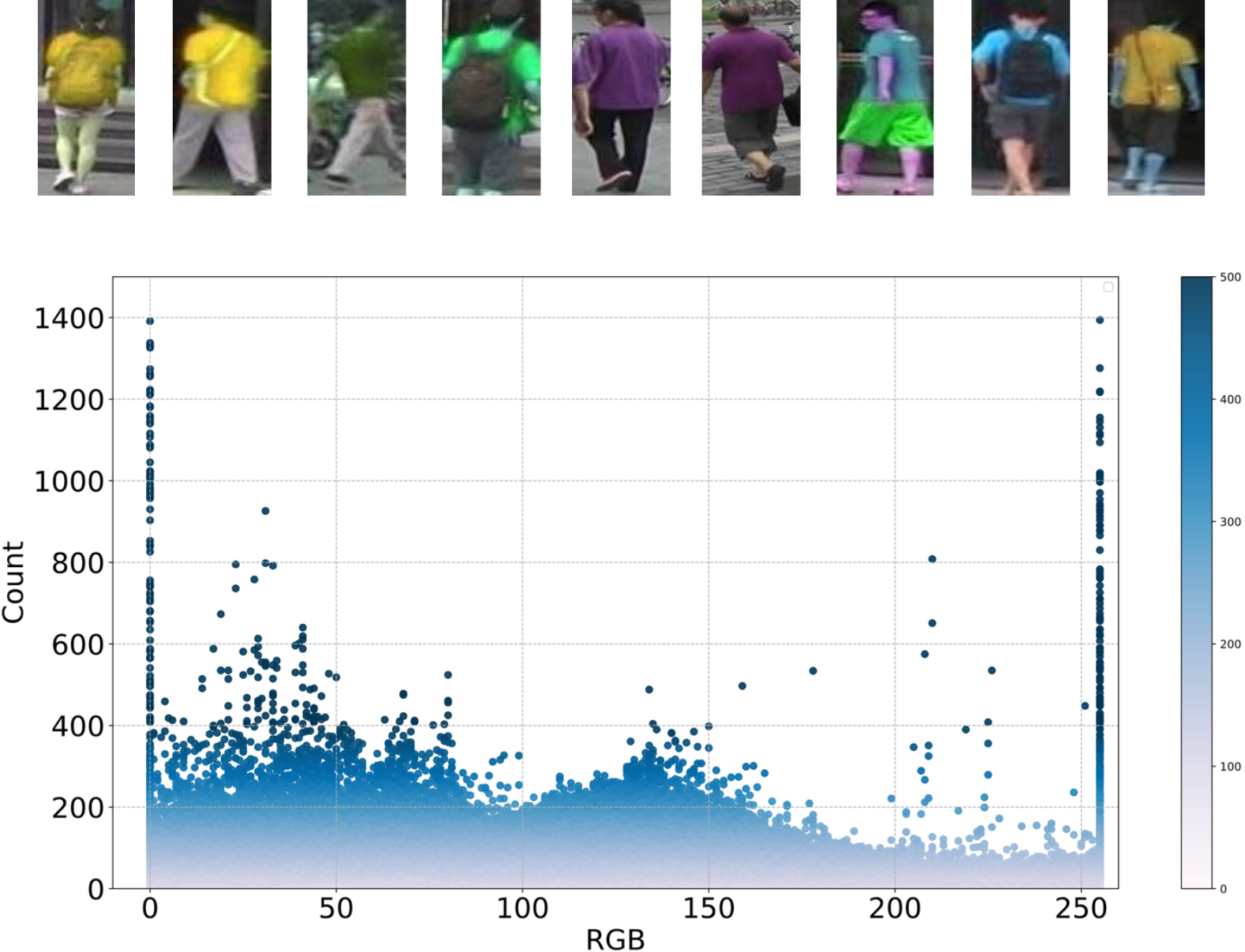}} 
\subfigure[Gray-Scale]{\includegraphics[trim={0cm 0cm 0cm 0cm},clip,width=0.67\columnwidth]{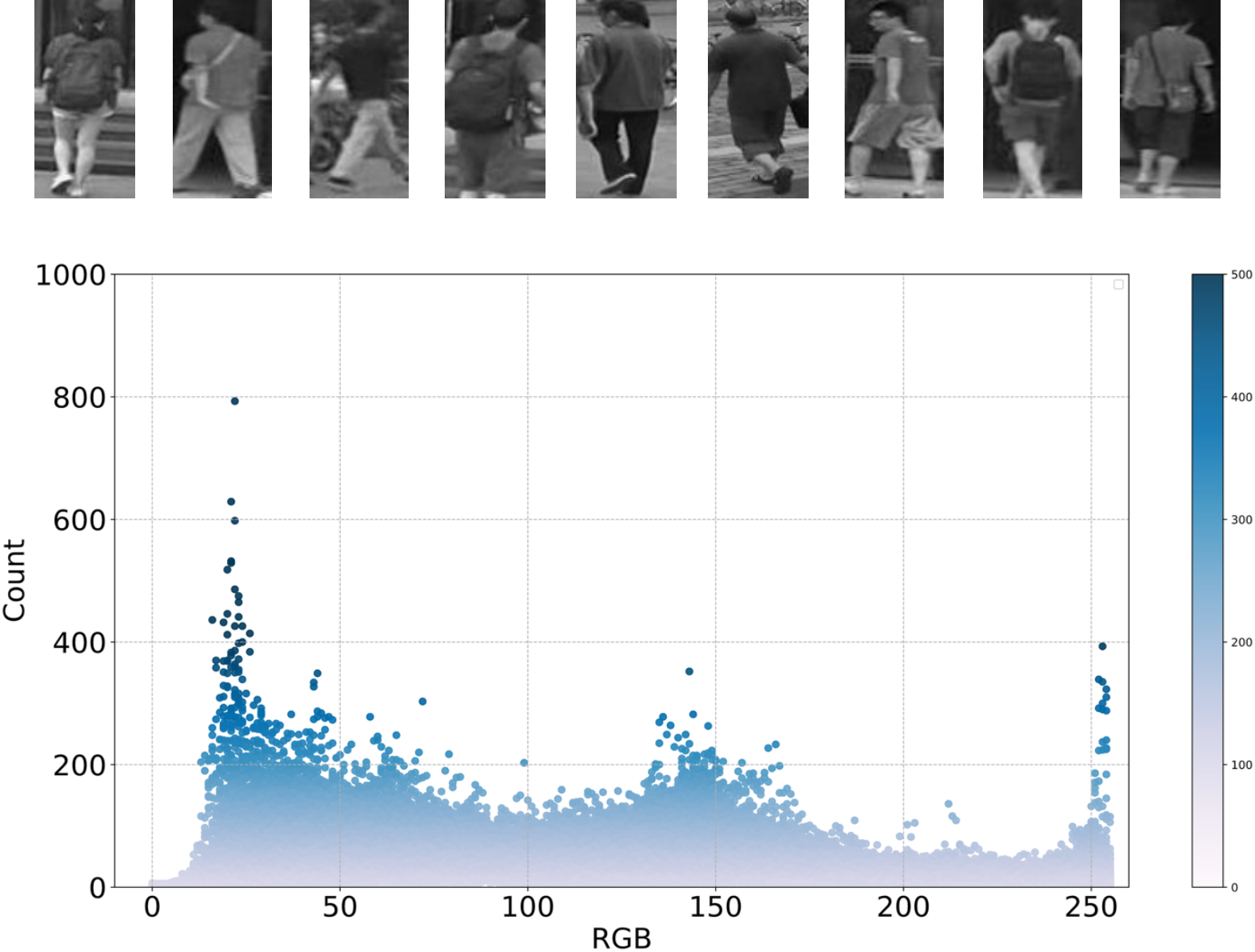}} 

% \subfigure[\hspace{-2pt}\footnotesize{$\G\circ\T^\prime$}]{\includegraphics[width=0.125\linewidth]{dataaug_3.pdf}} 

\end{center}
\caption{{  Comparison on distributions in histogram of intensity in RGB channels
under different data augmentation operations.}}
\label{fig:color_his}
\end{figure*}

Experimental results are provided in Table~\ref{Tab:different-color}.  We can read that
using ``color-jitter'' $\J(\cdot)$ yields some performance improvement, but using ``gray-scale'' $\G(\cdot)$ yield the best performance improvement.  
When combined with the cluster refinement step, we can observe the similar result that: using ``gray-scale'' $\G(\cdot)$ yields better performance improvement than using ``color-jitter'' $\J(\cdot)$. 
These results validate that suppressing colors is effective to gain performance improvement. % the model performance.
Compared to using ``gray-scale'', using ``color-jitter'' does not truly eliminate the influence brought by colors, that is to say, after using color-jitter, the color information still dominates.

%We account the reasons as the following two aspects: 
%a) compared to using gray-scale, using color-jitter does not truly eliminate the influence brought by colors, that is to say, after using color-jitter, the color information still dominates; and 
% b) color-jitter disturbs the color distribution in an unrealistic way, rather than consistently suppressing colors. 

\begin{table}
\small
\setlength{\tabcolsep}{1.2mm}{
\caption{{ Performance Comparison on using Color Data Augmentations and Gray-scale Transform to the Second Network Branch}.}
\begin{center}
\begin{tabular}{|c|c|c|c|c|c|}
					\hline		
					\multirow{2}{*}{Components} & \multirow{2}{*}{Cluster Refine}  & \multicolumn{4}{c|}{Market-1501}\\%&\multicolumn{4}{c|}{DukeMTMC-ReID}\\
					\cline{3-6}  
					& &mAP & Rank-1 &Rank-5&Rank-10\\
					
\hline\hline
 $\T^\prime(\cdot)$ &          & {70.3}&{87.4}& {94.6 }&{96.5}\\
%  $\J_s(\T^\prime(\cdot))$ &          &71.1 &87.4&93.9 &96.1\\
 $\J(\T^\prime(\cdot))$ &          &72.5 &87.8&95.3 &96.9\\
 $\G(\T^\prime(\cdot))$&           & 74.4 &89.3 &95.9 &96.7\\
 
 \hline
 
 $\T^\prime(\cdot)$ & \checkmark & 79.0 &90.6&  96.3 & 97.1\\
%  $\J_s(\T^\prime(\cdot))$ & \checkmark &78.9 &91.0&96.7 &97.7\\
 $\J(\T^\prime(\cdot))$  & \checkmark &79.1 &90.8 &96.7 &97.8\\
 $\G(\T^\prime(\cdot))$ & \checkmark &{80.9}&{92.7}& {97.4 }&{98.5}\\

\hline

\end{tabular}
\label{Tab:different-color}
\end{center}
}

\end{table}

To further reveal the mechanisms why using ``gray-scale'' works better than using ``color-jitter'' in the proposed framework, we show the statistic histograms of color distributions of using raw image, color-jitter, and gray-scale, respectively. 
Specifically, we compute the statistical histograms of the intensity values in the RGB channels of the raw color images and the images after using ``color-jitter'' and ``gray-scale'' with 500 images sampled at random in the training data from Market-1501. %as . %Specifically, we sample at random 500 images in the training data from Market-1501 to compute the statistical histograms. The %as 
The statistical results are shown in %Figure
Fig.~\ref{fig:color_his}.

%% check at Here May 2: 01:57//////////////////////////////////

We can observe that: using ``gray-scale'' yields roughly consistent distribution in the histogram compared to %as 
the raw images; whereas using the distribution in the histogram of the images after using ``color-jitter'' has some notable deviations from that of the raw images. In the histogram of using ``gray-scale'', the proportion of the pixels at the two extreme values (\ie, 0 and 255) are significantly reduced; whereas in the histogram of using ``color-jitter'', the proportion of the pixels at the two extreme values, especially at 0, are significantly magnified---this phenomenon might damage the content consistency with %in 
the raw image. The difference in the consistency of the histogram %distribution 
reveals the essential advantage of using ``gray-scale'' to suppress the influence of colors, rather than using ``color-jitter''. 

% This result shows that compared with using $\T^\prime(\cdot)$ , $\J_s(\cdot)$ provides a fixed change to color, which help is limited. Thus, what works significantly in $\J(\cdot)$ is the random changing. This unexpected change confuses the relation between color to identity to some extent. It can also be seen as a way to suppress color. But, from Figure~\ref{fig:color}, we can also find that compared to the gray-scale, color-jitter is still distinguishable in color. The random changes in images make a part of image confusion, and the influence of color difference still exists. This also limited the effectiveness of color-jitter and has worse performance than gray-scale $\G(\cdot)$. This results validate that suppressing colors is effective to improve the model performance.

\myparagraph{Evaluation on Parameters in DBSCAN} We conduct experiments 
evaluate the parameter $d$ to find the neighbors. In cluster refinement, we use DBSCAN with a smaller parameter $d^\prime$, where $d^\prime :=d-\delta$ to find the over-segmentation. 
%
%
%{\mc 
We conduct experiments on Market-1501 to evaluate the effects of changing the two parameters.
% }
%
Experiments are recorded in Table~\ref{Tab:cluster-parameter}. we can find
that while the change of $d$ will affect the baseline performance, our CACL  still improves the model performance significantly. Note that even though the baseline performance will sharply drop when using $d=0.7$, our method can also achieve a good performance which is also higher than other unsupervised methods in Table~\ref{tab:SOTA}.

The cluster refinement is  an important component in our proposed CACL, and $\delta$ is an important parameter %parameters 
to find the over-segmentation of the raw clusters. % step.  
Thus, we further conduct experiments to evaluate the performance of using different values of $\delta$. Experimental results are shown in 
Table~\ref{Tab:d and theta}. %Fig.~\ref{fig:smaller-eps}. 
We can find that the %model 
performance is not too sensitive to $\delta$. 
When using  $\delta = 0.02$, %the model achieves the best performance,
the performance achieves the best, \ie,  80.9/92.7\% at mAP/Rank-1 on Market-1501 and 69.6/82.6\% at mAP/Rank-1 on DukeMTMC-ReID.

%
%
%We can find that the model performance will be disastrous dropped. This proves that the model can not train with only the contrastive method in SimSiam~\cite{Chen:arxiv20}, and also proves that the cluster guidance is indispensable in CACL.

Moreover, we also test the stop-gradient operations under different structures. In %the 
Table~\ref{Tab:Ablation Study_inSupportingMaterial}, 
as can be read that, %we can find 
the performance of the framework with asymmetric structure drops slightly (\ie, only 0.7\% lower than that of using the stop-gradient operation) when the stop-gradient operation is not used. This hints that the framework with asymmetric structure in CACL does not highly depend on the stop-gradient operation.

\begin{table}
%\small 
\caption{Performance Comparison of different cluster parameter $d$ (the maximum  distance  between  neighbor points) on CACL and baseline method.}
\begin{center}
\setlength{\tabcolsep}{1.2mm}{
\begin{tabular}{|c|c|c|c|c|c|c|c|c|}
					\hline		
					\multirow{3}{*}{d} & \multicolumn{4}{c|}{Market-1501}&\multicolumn{4}{c|}{DukeMTMC-ReID}\\
					\cline{2-9}
					&\multicolumn{2}{c|}{Baseline}&\multicolumn{2}{c|}{CACL}&\multicolumn{2}{c|}{Baseline}&\multicolumn{2}{c|}{CACL}\\
					\cline{2-9}  
					& mAP & Rank-1  &mAP & Rank-1& mAP & Rank-1  &mAP & Rank-1\\
					
\hline\hline

0.4  &68.6& 85.9 &75.2&91.4&60.1&77.5&62.0&77.7\\
0.5  &71.2 &86.5 &{\bf81.6}&{\bf93.0}&63.4&80.3&67.5&81.8\\
0.6  &68.1 &85.2 &80.9&92.7&62.5&78.5&{\bf69.6}&{\bf82.6}\\
0.7  &43.8 &71.5 & 75.8&90.1&4.1&10.3 &66.7&80.6\\
\hline

\end{tabular}
}
\end{center}

\label{Tab:cluster-parameter}
\end{table}

\begin{table}
\caption{Illustration for the model performance with different  $\delta$ on Market-1501.}

\begin{center}
\setlength{\tabcolsep}{1.2mm}{
\begin{tabular}{|c|c|c|c|c|c|c|c|c|}
					\hline		
					\multirow{3}{*}{$\delta$} & \multicolumn{8}{c|}{Market-1501}\\
					\cline{2-9}
					&\multicolumn{2}{c|}{d = 0.4}&\multicolumn{2}{c|}{d = 0.5}&\multicolumn{2}{c|}{d = 0.6}&\multicolumn{2}{c|}{d = 0.7}\\
					\cline{2-9}  
					& mAP & Rank-1  &mAP & Rank-1& mAP & Rank-1  &mAP & Rank-1\\
					
\hline\hline

0.02  &75.2& 91.4 & 81.6&93.0&80.9&92.7&75.8&90.1\\
0.04  &70.8 & 89.5 &80.4& 92.6&80.3&92.3&68.7&86.2\\
0.06  &65.8 & 87.2 &77.7&91.7&79.0&91.4&8.20&20.3\\
0.08  &64.3 & 86.2 & 76.6&91.2&78.5& 91.3 &6.10&15.6\\
\hline

\end{tabular}
}
\end{center}

\label{Tab:d and theta}
% \vspace{-8pt}
\end{table}

% \begin{table}
% \small
% \setlength{\tabcolsep}{1.2mm}{
% \caption{Performance Comparison on using Color Data Augmentations and Gray-scale Transform to the Second Network Branch.}
% \begin{center}
% \begin{tabular}{|c|c|c|c|c|}
% 					\hline		
% 					\multirow{2}{*}{Components} & \multicolumn{4}{c|}{Market-1501}\\
% 					\cline{2-5}  
% 					& mAP & Rank-1 &Rank-5&Rank-10\\
					
% \hline\hline
% Data augmentation $\T^\prime(\cdot)$ & 79.0 &90.6&  96.3 & 97.1 \\
% Color-jitter $\J(\T^\prime(\cdot))$ &79.1 &90.8 &96.7 &97.8\\
% Gray-scale $\G(\T^\prime(\cdot))$  &80.9&92.7&97.4&98.5\\

% \hline

% \end{tabular}
% \label{Tab:different-color}
% \end{center}
% }

% \end{table}

% \begin{figure}
% \begin{center}
% %\fbox{\rule{0pt}{2in} \rule{.9\linewidth}{0pt}}
% \subfigure[$mAP$]{\includegraphics[width=0.9\linewidth]{figs/different-delta-map.pdf}}
% \subfigure[$Rank-1$]{\includegraphics[width=0.8\linewidth]{figs/different-delta-rank1.pdf}}
% \end{center}

% \caption{Illustration for the model performance with different  $\delta$ on Market-1501.} 
% \label{fig:smaller-eps}
% \vspace{-5pt}
% \end{figure}

\myparagraph{Evaluation Performance of Two Branches}
To further reveal the performance of the trained networks, we record the performance of using the output features of each branch of two networks $F(\cdot|\Theta)$ and $F^\prime(\cdot|\Theta^\prime)$, separately, %{ separately}, 
for person Re-ID in Table \ref{Tab:gray-intrain}. 
We can read that using the output features of the second branch $F^\prime(\cdot|\Theta^\prime)$ did yield significantly lower performance than that of using the output feature of the first branch $F(\cdot|\Theta)$, and the result of using $F^\prime(\cdot|\Theta^\prime)$ is similar to the result of the experiments without using $\mathcal{L}_C^{(intra)}$. 
This is because the second network branch pays attention to learning features from gray-scale images, lacking of the ability to capture richer information from color images.

\begin{table}
\small
\caption{Performance Comparison on $F(\cdot|\Theta)$ and $F^\prime(\cdot|\Theta^\prime)$.}
\begin{center}
\begin{tabular}{|c|c|c|c|c|}
					\hline		
					\multirow{2}{*}{Branch} & \multicolumn{4}{c|}{Market-1501}\\
					\cline{2-5}  
					& mAP & Rank-1 &Rank-5&Rank-10\\
					
\hline\hline

$F(\cdot|\Theta)$ (Color)  &80.9&92.7&97.4&98.5\\
$F^\prime(\cdot|\Theta^\prime)$ (Gray-Scale)  &43.8 &71.5 &83.9 &87.1\\
\hline

\end{tabular}
\end{center}

\label{Tab:gray-intrain}
% \vspace{-8pt}
\end{table}

\begin{table}
\small
\caption{Performance Comparison on different $\alpha$.}
\begin{center}
\begin{tabular}{|c|c|c|c|c|}
					\hline		
					\multirow{2}{*}{Branch} & \multicolumn{4}{c|}{Market-1501}\\
					\cline{2-5}  
					& mAP & Rank-1 &Rank-5&Rank-10\\
					
\hline\hline

0.0  &75.1&89.8&96.3&97.3\\
0.2   &{\bf80.9}&{\bf92.7}&{\bf97.4}&{\bf98.5}\\
0.4  &80.8 &92.5 &97.1 &98.2\\
0.6  &80.2 &92.4 &97.2 &98.3\\
0.8  &77.3 &90.9 &96.6 &98.0\\
1.0  &4.3 &10.9 &19.9 &24.9\\
\hline

\end{tabular}
\end{center}

\label{Tab:momentum}
% \vspace{-8pt}
\end{table}

\myparagraph{Evaluation on Memory Update Parameter $\alpha$}  We conduct experiments to evaluate the effects %iveness 
of the memory update parameter $\alpha$ and show the results in Table~\ref{Tab:momentum}. We can find that our CACL % the method 
is not sensitive to the changing of memory update parameter $\alpha$, except for $\alpha=1$. When using $\alpha=1$, the model performance significantly drops %ped 
because the memory bank has %will 
not been updated at this time. When using $\alpha=0.2$ the model achieves the best performance on Market-1501, \ie,  80.9/92.7\% at mAP/Rank-1.

% Groundt
\myparagraph{Evaluation on Performance with Ground-truth Labels}  We compare %test both 
our CACL %in unsupervised setting 
to %and 
the baseline method with the ground-truth %supervised 
labels (\ie, in supervised setting). The results are %is 
shown in Table~\ref{Tab:orac}. We can find that CACL could achieve good performance under unsupervised setting, which is merely lower 3/1.1\% at mAP/Rank-1 than the baseline method, which is trained with the ground-truth labels on Market-1501. %What's more, 
Moreover, if we provide ground-truth labels to train our CACL (\ie, CACL+labels), notable improvements in performance than the supervised baseline method can be observed. 
% improve the model performance under a supervised setting, which 1.8/0.6\% mAP/Rank-1 on Market-1501 and 1.6/0.6\% mAP/Rank-1 on DukeMTMC-ReID.

\begin{table}
\begin{center}
\caption{%Performance Comparison of ground truth label and pseudo label, where Oracle* means that we use the ground-truth labels to train the baseline method and Oracle means that we use the ground-truth labels to train our CACL.
Performance Comparison to Baseline Method in Supervised Setting. ``baseline + labels'' means that we use the ground-truth labels to train the baseline method; whereas ``CACL + labels'' means that we use the ground-truth labels to train our CACL.
}
\begin{tabular}{|c|c|c|c|c|}
					\hline		
					\multirow{2}{*}{Method} & \multicolumn{2}{c|}{Market-1501}&\multicolumn{2}{c|}{DukeMTMC-ReID}\\
					\cline{2-5}  
					& mAP & Rank-1  & mAP & Rank-1 \\
					
\hline\hline

%Oracle*  &83.9& 93.6&73.3& 86.6\\
%Oracle  &{\bf 85.7 }&{\bf 94.2 }&{\bf74.9}&{\bf87.2}\\

CACL                &{80.9}&{92.7}&{69.6}&{ {82.6}}\\
Baseline + labels   &83.9& 93.6&73.3& 86.6\\
CACL + labels       &{\bf 85.7 }&{\bf 94.2 }&{\bf74.9}&{\bf87.2}\\
\hline

\end{tabular}
\label{Tab:orac}
\end{center}

% \vspace{-8pt}
\end{table}

\begin{figure*}[!ht]
\small
\begin{center}
%\fbox{\rule{0pt}{2in} \rule{.9\linewidth}{0pt}}
\includegraphics[width=0.75\linewidth]{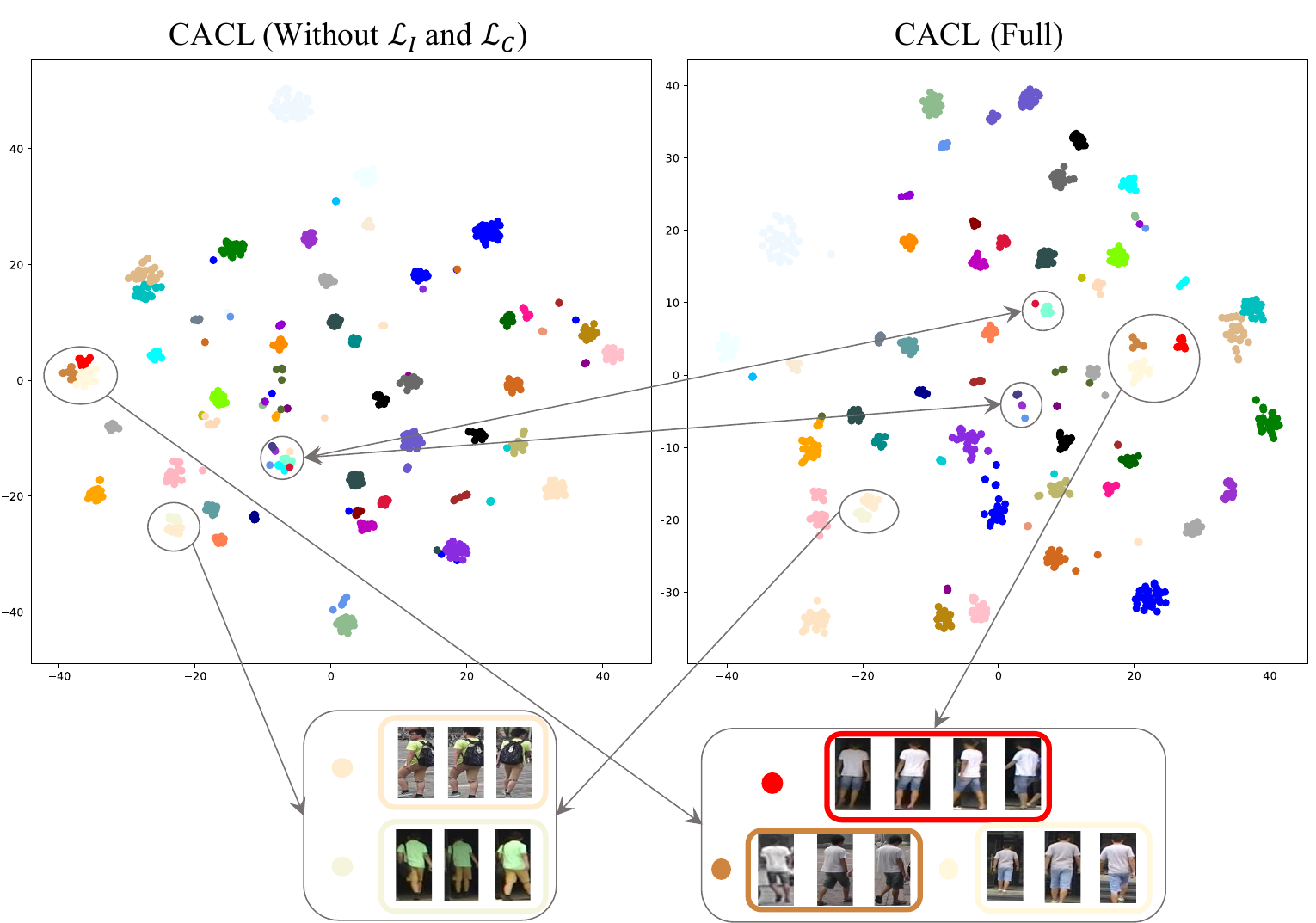} % flowchart.pdf
\end{center}
  \caption{%The T-SNE 
  Data Visualization via $t$-SNE of the learned feature and clusters under two different training strategies: Training without $\mathcal{L}_C$ and $\mathcal{L}_I$ (left) as mentioned in 
  Table~\ref{TAb:ab} and our CACL (right). The data points come from the Market-1501 training set (1,000 images of 60 identities). The points with the same color mean the image of the same identity. To demonstrate the difference between the two distributions in detail, we further zoom in on the circled clusters and show the corresponding images.   
  The images in the boxes are similar to each other and the corresponding data points are very close to each other or even overlapping in the feature space if the model is trained without using $\mathcal{L}_C$ and $\mathcal{L}_I$, as shown in the left box; whereas using the contrastive losses $\mathcal{L}_C$ and $\mathcal{L}_I$ will %efficient
  effectively distinguish these data points and maintain the cluster compactness as shown in the right box. }
  
\label{fig:feature_contra}
\end{figure*}

\begin{figure*}[!ht]
\small
\begin{center}
%\fbox{\rule{0pt}{2in} \rule{.9\linewidth}{0pt}}
\includegraphics[width=1.0\linewidth]{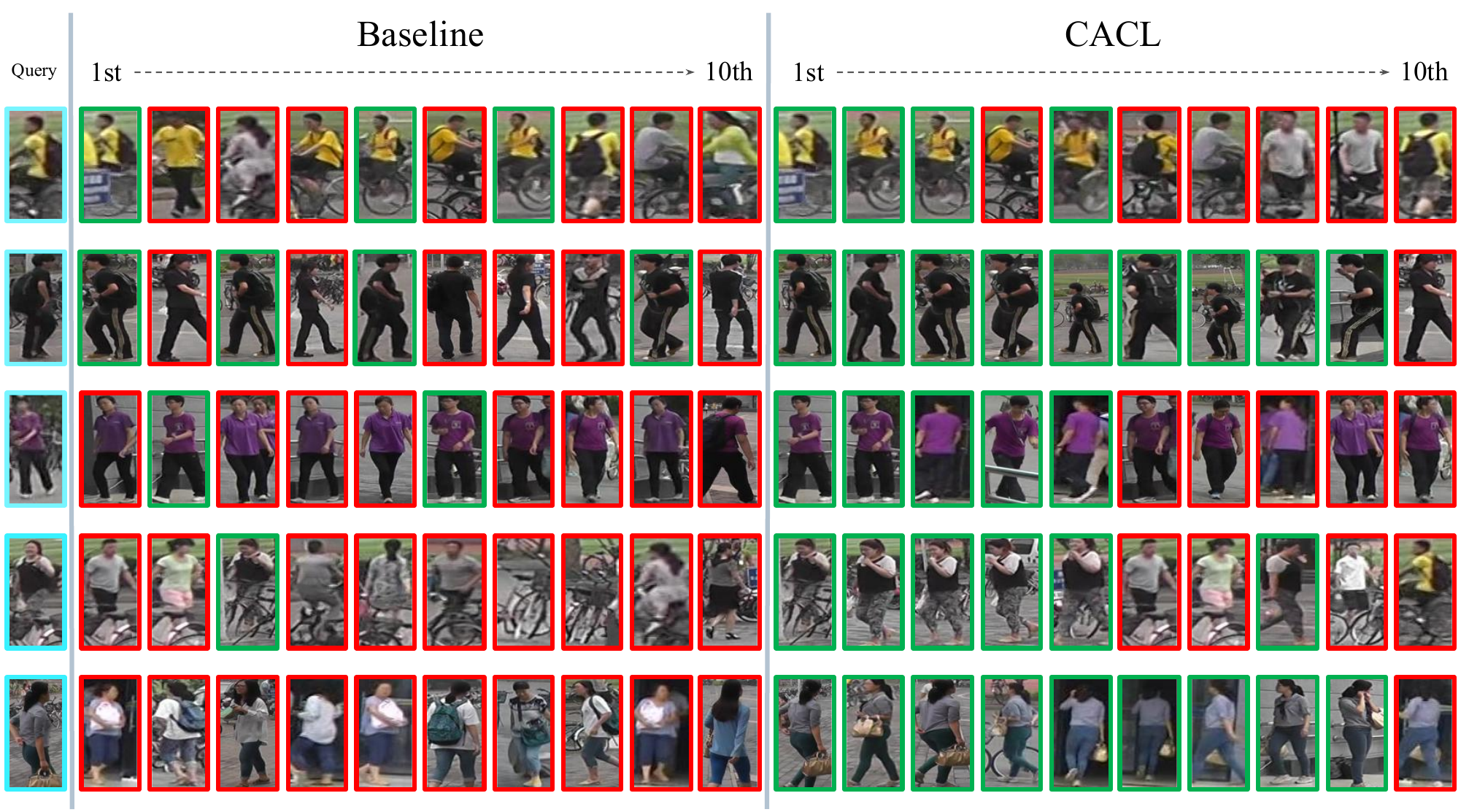} % flowchart.pdf
\end{center}
  \caption{%The T-SNE 
  Visualization of the top-10 best matched images. We show the top-10 best matching samples in the gallery set for the query sample with the baseline method and our proposed CACL. The images with frames in %blue, 
  green and in red are % query images, 
  the correctly matched images and %error 
  mismatched images, respectively.}
  
\label{fig:retivel}
\end{figure*}

\subsection{Data Visualization} 
To gain some intuitive understanding of the performance of our proposed CACL, we conduct a set of data visualization experiments on Market-1501 
to visualize the clustering results of the learned features when different training strategies are used: a) without using the contrastive loss % $\mathcal{L}_C$ and $\mathcal{L}_I$; 
$\mathcal{L}_C + \mathcal{L}_I$; and b) using the contrastive losses $\mathcal{L}_C + \mathcal{L}_I$. 

Experimental results are shown in Fig.~\ref{fig:feature_contra}. We can observe that the contrastive loss %$\mathcal{L}_C$ and $\mathcal{L}_I$ 
$\mathcal{L}_C + \mathcal{L}_I$ did help the model distinguish those similar images while maintaining the cluster compactness, and also separate the overlapping individual samples from each other. 
This confirms the effectiveness of our proposed approach, and it also shows that our approach can attenuate the influence of clothing color.

At the same time, we also selected some query samples with the top-10 %ten % 
best matching images in the gallery set and show them in Fig.~\ref{fig:retivel}. %As shown in Fig.~\ref{fig:retivel}. 
Compared to the baseline model, our approach %method 
returns more accurate results. We can find that most of the wrong samples matched by the baseline model are dressed in the same color with %as 
the query sample. These results suggest that our approach can effectively ignore the interference caused by samples with similar colors and thus find more accurate matches.

\section{Conclusion}
\label{sec:conclusion}

%% cluster structure --> cluster information or cluster result

We have proposed a Cluster-guided Asymmetric Contrastive Learning (CACL) approach for unsupervised person Re-ID, in which cluster information %structure 
is leveraged to guide the feature learning in a properly designed contrastive learning framework. Specifically, in our proposed CACL, instance-level contrastive learning is conducted with respect to the asymmetric data augmentation and cluster-level contrastive learning is conducted with respect to the refined clustering result. %structure. 
By leveraging the refined cluster result into contrastive learning, CACL is able to effectively exploit the invariance within and between different data augmentation views for learning more effective features beyond the dominating colors. In addition, we confirmed that refined clustering result %this paper also proposes a clustering refinement method, which 
could help our CACL approach mine invariant information more effectively at the cluster level. 
We have conducted extensive experiments on three benchmark datasets and demonstrated the superior performance of our proposal. 

%As the future work, it is promising to incorporate attention mechanism to enhance the feature representation, \eg, \cite{Si:CVPR18}\cite{。。。ViT}, re-ranking to improve the overall performance \cite{}, clustering ensemble or hybrid contrastive learning strategy, \eg, \cite{Sun:ACPR21}

As the future work, it is interesting and promising to incorporate attention mechanism (\eg, \cite{Si:CVPR2018, VIT:ICLR2021}), clustering ensemble and hybrid contrastive learning strategy (\eg, \cite{Sun:ACPR21}) or side information in dataset (\eg, \cite{chen:cvpr2021})  to further enrich the representation capacity, improve the stability and enhance the overall performance of the proposed framework. 
% Moreover, it is also worth to explore whether suppressing dominating color helps to improve the performance of vehicle re-identification (\eg, \cite{liu:IEM2017}).} 
{ What's more, in other related fields, such as face recognition or vehicle re identification (\eg, \cite{liu:ICME2016,liu:IEM2017}), whether suppresses the dominating color can also  bring positive influence is a very interesting and worth exploring direction.}

\ifCLASSOPTIONcaptionsoff
  \newpage
\fi



%\bibliographystyle{IEEEtran}

{

\bibliographystyle{IEEEtran}
\bibliography{biblio//mkli_ReID,biblio//cgli,biblio//cqlin_gcn_reid}

}

\end{document}